\documentclass[preprint,authoryear,3p]{elsarticle}
\usepackage{amssymb,amsthm,amsmath}
\usepackage{graphicx}
\usepackage{subfig}
\usepackage{float}
\usepackage{multirow}
\usepackage{longtable,lscape}
\usepackage{cases}
\usepackage{url}
\usepackage{caption}
\usepackage{color}
\usepackage{amsthm}
\usepackage{algpseudocode,algorithm,algorithmicx}
\usepackage[pagewise]{lineno}
\usepackage{comment}
\usepackage{graphicx}
\usepackage{rotating}
\usepackage{bbm}

\usepackage{diagbox}

\usepackage{enumitem}
\usepackage{fancyhdr}
\usepackage{fullpage}
\usepackage{makecell}

\theoremstyle {plain}% default

\newcounter{x}

\theoremstyle{definition}
\newtheorem{defn}{Definition}[section]

\newtheorem{exmp}{Example}[section]

\theoremstyle{remark}
\newtheorem*{rem}{Remark}

\usepackage{titlesec}
\titleformat{\paragraph}
{\normalfont\normalsize}{\theparagraph}{1em}{}
\titlespacing*{\paragraph}
{0pt}{3.25ex plus 1ex minus .2ex}{1.5ex plus .2ex}

\usepackage{color}

\usepackage{lipsum}

\def\ps@pprintTitle{%
	\let\@oddhead\@empty
	\let\@evenhead\@empty
	\def\@oddfoot{\reset@font\hfil\thepage\hfil}
	\let\@evenfoot\@oddfoot
}
\makeatother

\makeatletter
\def\ps@pprintTitle{%
	\let\@oddhead\@empty
	\let\@evenhead\@empty
	\let\@oddfoot\@empty
	\let\@evenfoot\@oddfoot
}
\makeatother

\usepackage{multicol}

\journal{Transportation Research Part C}

\begin{document}

%\linenumbers

\begin{frontmatter} 
		

\noindent
\textcolor{red}{
Published in: Transportation Research Part C 137 (2022) 103560.}

\noindent
\textcolor{blue}{
This preprint is originally revised and extended from its previous version (https://arxiv.org/abs/2011.10915v1), so there may exist some overlaps. In the previous version, we proposed a framework of Markov routing games (MRG) and introduced the bi-level network design problems. In the latest one, we first modify the MRG framework by considering the linkage between MRG and dynamic traffic assignment (DTA). We then incorporate various DTA models into MRG and investigate game equilibria for agents. We remove the bi-level network design problems from the previous version. Due to significant differences by major revision, it is recommended to read and make use of this latest version. The previous version shall be removed.}

\noindent
\textcolor{blue}{
Please cite this Paper as: Shou, Z., Chen, X., Fu, Y., Di, X., 2022. Multi-Agent Reinforcement Learning for Markov Routing Games: A New Modeling Paradigm For Dynamic Traffic Assignment. Transportation Research Part C: Emerging Technologies 137, 103560. doi:https://doi.org/10.1016/j.trc.2022.103560}

\title{
Multi-Agent Reinforcement Learning for Markov Routing Games: A New Modeling Paradigm For Dynamic Traffic Assignment
%And Its Linkage To Dynamic User Equilibrium
%: \\
%A Unified Paradigm %of Model-Based and Data-Driven Approaches
% Bilevel Optimization for En-Route Path Choice 
% in Congestion Games 
% Using Multi-Agent Reinforcement Learning
}
		
\date{\today}
		
		
\author[cu]{Zhenyu Shou}
\author[cu]{Xu Chen}
\author[cu]{Yongjie Fu}

\author[cu,dsi]{Xuan Di\corref{cor}}
\ead{sharon.di@columbia.edu}

\cortext[cor]{Corresponding author. Tel.: +1 212 853 0435;}

\address[cu]{Department of Civil Engineering and Engineering Mechanics, Columbia University}
\address[dsi]{Data Science Institute, Columbia University}

\begin{abstract}
    This paper aims to develop a paradigm that models the learning behavior of intelligent agents (including but not limited to autonomous vehicles, connected and automated vehicles, or human-driven vehicles with intelligent navigation systems where human drivers follow the navigation instructions completely) with a utility-optimizing goal and the system's equilibrating processes in a routing game among atomic selfish agents. 
    Such a paradigm can assist policymakers in devising optimal operational and planning countermeasures under both normal and abnormal circumstances. 
    To this end, we develop a Markov routing game (MRG) in which each agent learns and updates her own en-route path choice policy while interacting with others in transportation networks. 
    %We also illustrate that the equilibrium outcomes computed from our developed MARL paradigm coincide with DUE. %\tcr{and dynamic system optimal (DSO), respectively}, 
    %when rewards are set differently. 
    %With the prevalence of navigation devices, people make route choices according to downstream traffic information, which is not observable previously without information systems. We aim to understand how information structure plays a role in travelers' learning behavior and equilibrating dynamics. 
	%Traffic congestion has become one of the major issues faced by modern cities. It is therefore of great importance to research into traffic assignment problems. 
	%This paper aims to tackle the multi-driver route choice task by 
	%Noticing that drivers are not the only player in a route choice task, city planners can directly impact route choice behavior of drivers by imposing some control (e.g., increasing or decreasing the capacity of links in a transportation network). 
	To efficiently solve MRG, we formulate it as multi-agent reinforcement learning (MARL) and devise a mean field multi-agent deep Q learning (MF-MA-DQL) approach that captures the competition among agents. 
	The linkage between the classical DUE paradigm and our proposed Markov routing game (MRG) is discussed. We show that the routing behavior of intelligent agents is shown to converge to the classical notion of predictive dynamic user equilibrium (DUE) when traffic environments are simulated using dynamic loading models (DNL). 
	In other words, the MRG depicts DUEs assuming perfect information and deterministic environments propagated by DNL models. Four examples are solved to illustrate the algorithm efficiency and consistency between DUE and the MRG equilibrium, on a simple network without and with spillback, the Ortuzar Willumsen (OW) Network, and a real-world network near Columbia University's campus in Manhattan of New York City.


\end{abstract}
		
\end{frontmatter}

\section{Introduction}
\label{sec:intro}

%\textcolor{red}{R1.2: Introduction, "With a growing number of drivers relying on GoogleMaps or other navigation tools for dynamic routing, one question of interest is that, when everybody follows the advisory of shortest paths provided by one central platform, will these drivers still be able to gain from taking these paths?" I cannot find the result to directly relate to this question.}
%With a growing number of drivers relying on GoogleMaps or other navigation tools for dynamic routing, one question of interest is that, when everybody follows the advisory of shortest paths provided by one central platform, will these drivers still be able to gain from taking these paths? 

%This paper aims to tackle this question by accounting for competition among drivers. 
When one chooses a path dynamically while navigating a road network, others may also do so to compete for limited road resources. 
Therefore it is crucial to model the dynamic routing choices of many drivers simultaneously. 
This paper aims to develop a game-theoretic paradigm that models one's learning behavior and the system's equilibrating processes in a dynamic routing game among atomic selfish agents, while accounting for competition among drivers. 
%To address the above problem, we will first review the literature on route choice of single-agent and then move to that of multi-agent. 
% \tcb{
% \begin{rem}
% \begin{enumerate}
%     \item In this paper, we assume travelers are perfectly rational agents in the context of selfish routing. In other words, every traveler is a self-interest agent who aims to maximize individual accumulative rewards or minimize individual cost or maximize payoffs. 
%     \item We use agent, player, and controllable cars interchangeably, which all represent those intelligent vehicles that select routes based on utility-optimizing behavior. 
% \end{enumerate}
% \end{rem}
% }

%\subsection{Single-agent route choice}

While traveling from an origin to a destination, one needs to select a sequence of road segments. Thus, en-route path choice is inherently a sequential decision-making process, in which at each junction one decides what the next link to take. 
Markov decision processes (MDP) \citep{puterman_markov_1994} and reinforcement learning (RL) \citep{sutton_introduction_1998} have become popular tools to model such a sequential process, in which travelers are rational agents to optimize a prescribed objective function (or reward) \citep{seongmoon_kim_optimal_2005, kim_solving_2016}.
%The assumption of rationality has been challenged by various behavioral and economic theories, including, to name a few, bounded rationality \citep{di_boundedly_2013,di_boundedly_2016}, prospect theory \citep{gao2010adaptive,xu2011prospect,yang2014development,zhang2018cumulative}, and regret theory \citep{de2011expected}. 
In literature, recently there is a growing body of research using RL based approaches on route choice behavior. 
We will first review the literature on route choice of single-agent and then move to that of multi-agent. 

\subsection{Literature on reinforcement learning based route choices}

The existing literature on dynamic routing games using RL-based approaches is listed in Table~\ref{tab:marl}.
Based on how to define the action of an agent, we broadly categorize RL-based route choice models into two groups. In the first group, the action of an agent is a route \citep{zhou_reinforcement_2020, ramos_analysing_2018, stefanello_using_2016}. For example, in \cite{ramos_analysing_2018, stefanello_using_2016}, for an agent aiming to go to her destination node from her origin node, the action space for her is the $k$ shortest paths from her origin to her destination. These studies assume that every driver does follow her chosen route until she reaches her destination. In reality, however, drivers could deviate from their chosen route when they realize that traffic condition on some alternative routes might be better, especially when they get stuck in traffic on the current route. To capture this en-route adaptive behavior of drivers, the second group of studies focus on en-route traffic assignment, or route choice from the perspective of a driver. In these studies, the action space of an agent currently at a node is the outbound links from the node \citep{grunitzki_individual_2014, bazzan_multiagent_2016, mao_reinforcement_2018}. In other words, every agent needs to decide which outbound link to choose whenever the agent arrives at a node, until the agent reaches some terminal node. This en-route decision-making process captures the adaptive behavior of real drivers. 

%\textcolor{blue}{(Based on current structure, maybe move single-agent RL to 1.1?)}
To study the en-route decision-making process of drivers, both single-agent RL and multi-agent RL are used in the literature. A single-agent Q-learning approach is developed in \cite{mao_reinforcement_2018} with the use of some global information such as congestion status in the definition of state. The authors bootstrapped two traffic profiles from the PeMS dataset and derived an optimal policy for the agent. Unfortunately, single-agent RL fails to capture the competition among adaptive agents. Multi-agent RL is recently used to tackle the multi-driver route choice task \citep{grunitzki_individual_2014, bazzan_multiagent_2016}. 
Despite of modeling multi-driver interactions, these studies use independent multi-agent tabular Q-learning where every agent is treated as an independent learner who has no information of other agents. %\textcolor{blue}{What's independent Q-learning? not clear here.}
The assumption of independent learners suffers from two issues. 
First, it is impractical to assume that agents select routes independently without accounting for interactions with other agents and congestion effects on transportation networks.
%Another key issue arises when independent learners treat other agents as part of the stochastic environment, that is, 
Second, theoretical convergence guarantee for Q-learning does not hold, because the environment is no longer Markovian and stationary \citep{matignon_independent_2012, nguyen_deep_2020}.

To fill the aforementioned research gaps, 
in this paper, we will model en-routing decision-making processes in a multi-agent system as a Markov game in which one's payoff function depends on all others' routing strategies. 

\subsection{MARL and multi-agent Markov game}
As a game-theoretic framework for MARL, Markov game \citep{littman_markov_1994,solan2015stochastic} is a generalization of MDP-like environments to multiple interacting agents with competing goals, in which the environment makes transitions probabilistically in response to the agents' actions. Markov game is typically defined as a non-cooperative game where self-interested agents aiming to maximize their own payoffs.  In a Markov game, the solution concept of Nash equilibrium is generally adopted \citep{hu_multiagent_1998}. At a Nash equilibrium, one agent's strategy (i.e., policy) is the best response to other agents' strategy. MARL is an efficient and versatile tool to solve for the optimal policy of each agent. Among the pioneering studies, \cite{littman_markov_1994} proposes a minimax Q-learning algorithm to solve for the optimal policies of two agents with opposed goals in a zero-sum game, and \cite{hu_multiagent_1998} extends it to a general-sum game and provided a multi-agent Q-learning algorithm. 

Recent literature has witnessed various applications of MARL to high-dimensional and complicated tasks such as playing the game of Go \citep{silver_mastering_2016,silver_mastering_2017}, Poker \citep{brown_superhuman_2018, brown_superhuman_2019}, Dota 2 \citep{OpenAI_dota}, and StarCraft II \citep{vinyals_grandmaster_2019}. Outside the computer game domain, MARL has also attracted significant attention and been used in energy sharing \citep{prasad_multi-agent_2019}, federated control \citep{kumar_federated_2017}, and sequential social dilemma \citep{leibo_multi-agent_2017}, to name a few. 
Interested readers are referred to \cite{nguyen_deep_2020} for a comprehensive review of applications and challenges of MARL. 
In the transportation domain, we have also seen a growing trend of applying MARL to vehicle routing \citep{bazzan_re-routing_2008, bazzan_multiagent_2016}, traffic signal control \citep{bakker_traffic_2010,chen_toward_2020}, fleet management \citep{lin_efficient_2018, shou_optimal_2020,shou_reward_2020,di2018link,di2019rue_e,chen2021RUE}, order dispatching \citep{li_efficient_2019}, and autonomous driving \citep{palanisamy_multi-agent_2019, bhalla_deep_2020}. 
However, its application to dynamic routing game is still at its nascent stage.

\begin{sidewaystable}%
	\centering\caption{Existing research on dynamic routing using RL based approaches}
	\label{tab:marl}
	%\begin{tabular}{p{50pt} p{60pt}  p{35pt} p{80pt} p{110pt} p{30pt} p{60pt}} 
	\begin{tabular}{p{50pt}  p{75pt} p{30pt} p{55pt} p{80pt} p{80pt} p{60pt} p{35pt} p{120pt}} 
		\hline \hline
		%\multicolumn{3}{|c}{\multirow{2}{*}{}} & \multicolumn{2}{|c|}{Driver 1} \\ \cline{4-5}
		%& & & $\#1$ & $\#4$ \\ \cline{1-5}
		%\multirow{2}{*}{Driver 2} & \multicolumn{2}{|l|}{$\#1$} & $1.5, 1.5$ (Figure~\ref{subfig:case4})& $7,3$ (Figure~\ref{subfig:case2})\\ \cline{2-5}
		%& \multicolumn{2}{|l|}{$\#4$} & $3,7$ (Figure~\ref{subfig:case3})& $3.5,3.5$ (Figure~\ref{subfig:case1})\\ \hline
		 Research type  & Reference & Setting  & State & Action set & Reward  & Algorithm & Bilevel optimization & Gap\\ \hline
		 \multirow{3}{*}{Route-based} & \cite{stefanello_using_2016}  & Multi-agent &  - & $k$ shortest routes from origin to destination &  Negative travel time & Independent tabular Q-learning & No & - \\ 
		 & \cite{ramos_analysing_2018} & Multi-agent  & - & $k$ shortest routes from origin to destination & App-based regret (anticipated disutility) & Independent tabular Q-learning & No & - \\ 
		  &\cite{zhou_reinforcement_2020} & Multi-agent & - & Feasible routes from origin to destination & Negative travel time & Bush-Mosteller RL scheme  & No & -\\ \hline
		 \multirow{4}{*}{En-route}  %& \cite{bazzan_re-routing_2008}  & & &  & &  &  &\\
	     & \cite{grunitzki_individual_2014} &  Multi-agent & (node)  & Outbound links  & Negative travel time (Individual and systematic) & Independent tabular Q-learning & No & \multirow{1}{\linewidth}{Independent learning suffers from non-stationary and non-Markovian issues; tabular Q-learning could not handle continuous or large state/action space}\\
		 & \cite{bazzan_multiagent_2016} & Multi-agent &  (node) & Outbound links & Negative travel time & Independent tabular Q-learning  & No & \\
		 & \cite{mao_reinforcement_2018}& Single-agent & (time, node, congestion status) & Successor nodes  & Negative travel time &  Tabular Q-learning & No & Using global information, i.e., congestion status in both training and execution\\ 
		 & This study & Multi-agent & (time, node) & Outbound links & Negative travel cost & Mean-field deep Q-learning & Yes &\\ \hline
		\hline
	\end{tabular}
\end{sidewaystable} %

\subsection{Contributions of this paper}

% Noticing that in a road network, in addition to the infrastructure supply (i.e., road resources) and travelers (i.e., traffic demand), there is another player, that is, city planner, who can impact the behavior of travelers.% by imposing some administrative measures such as tolling and signal control. %We note that increasing or reducing the infrastructure supply could be regard
% With selfish and rational travelers aiming to minimize their travel cost, city planners can achieve some systematic goal by imposing externalities on individuals' cost via various operational measures such as congestion pricing or traffic signal control. %taking advantage of this selfishness of travelers. 
% %For example, imposing a proper toll charge on some links may reduce the overall travel cost of all travelers. %We will examine the effect of toll charge and signal control in Section~\ref{sec:bilevel} and \ref{sec:case}, respectively. 
% Therefore, we formulate such interactions between travelers and city planners as a bilevel optimization task. We employ the developed flow-dependent multi-agent deep Q learning approach on the lower level to solve for optimal route choices of travelers and a Bayesian optimization scheme on the upper level to solve for optimal controls by city planners. 

    %We have briefly discussed the advantage of MARL over DTA models, which is a radical contribution of this paper. 
    Here we will highlight the main contributions of this study as follows. 
    
\begin{enumerate}
    \item Multi-agent dynamic routing is modeled as a Markov routing game (MRG) that accounts for the interactions among adaptive agents and traffic congestion. It is capable of modeling individual agents' routing behavior with a partial or unknown information structure and a model-free environment, in which agents need to learn optimal policies through repeated interactions with other agents and the traffic environment.
    
    \item %First, 
    %\tcr{a flow-dependent mean field deep Q-learning algorithm is developed to tackle the route choice task of multi-commodity multi-class agents. The flow-dependent mean action not only partially captures the competition among agents but also enables Q-value sharing and policy sharing. }
    
    An efficient multi-agent reinforcement learning algorithm is developed %that captures the interaction among adaptive agents 
    via a mean action, which is defined as the traffic flow on the chosen link right after an agent enters the link. The use of the mean action not only captures the competition among agents, but also enables value sharing and policy sharing, which is computationally favorable, especially in a large-sized problem with a large number of agents. 
%we develop a flow-dependent multi-agent deep Q-learning approach which captures the interaction among adaptive agents via a flow-dependent mean action. The flow-dependent mean action is defined as the traffic flow on the chosen link right after an agent enters the link. The use of the flow-dependent mean action not only captures the competition among agents, but also enables Q-value sharing and policy sharing, which is computationally favorable, especially in a large-sized problem with multi-commodity multi-class agents. 

    \item  %Second, 
    A variety of dynamic network loading (DNL) models are used as the simulation environment of the proposed Markov routing game. We then demonstrate the consistency of our computed Markov equilibrium and its associated dynamic user equilibrium (DUE). 
    The linkage between the classical dynamic traffic assginment (DTA) paradigm and our proposed game is also demonstrated. We illustrate that the developed Markov game for dynamic routing depicts analytical DTA assuming complete information of environment transition dynamics and travel or delay time functional forms. %and deterministic environments propagated by DNL models.
    %This paper is the first-of-its-kind to unify the model-based (i.e., DUE) and data-driven (i.e., MARL) paradigms for dynamic routing games. 
    
    \item %Third, 
    % we formulate the overall task including travelers and city planners as a bilevel optimization task. We demonstrate the effect of two administrative measures, namely tolling and signal control, on the behavior of travelers and show that the systematic objective of city planners can be optimized by a proper control. 
    Computational efficiency of the developed Markov routing game is demonstrated on mid- and large-sized networks.
\end{enumerate}

	The remainder of the paper is organized as follows. Section~\ref{sec:mdmarl} introduces the developed Markov game for dynamic routing and devises a mean field multi-agent deep Q-learning algorithm. 
% 	Section~\ref{sec:mdmarl} introduces the developed flow-dependent mean field multi-agent deep Q-learning algorithm. 
% 	We detail the necessity and benefits of using the flow-dependent mean action which carries not full but partial information of nearby agents.
	Section~\ref{sec:DUE} demonstrates the linkage between the classical DUE paradigm and our proposed MARL paradigm. Section~\ref{sec:case} presents four examples to demonstrate the consistency between DUE and the equilibrium learned from our model as well as algorithm efficiency, on a simple network without and with spillback, the OW network, and a real-world network near Columbia University's campus in Manhattan of New York City.
	%Section~\ref{sec:bilevel} introduces the bilevel optimization model where the upper level city planners can impact the behavior of lower level travelers through operational measures. A case study of the application of the bilevel optimization model to the Braess network with tolling is presented. 
	%\tcr{Section~\ref{sec:case} presents the application of the bilevel optimization model to solve for optimal traffic signal control over a real-world large-scale road network near Columbia University's campus in the City of New York.}
	Section~\ref{sec:conclu} concludes this study.

\section{From singe- to multi-agent routing models} \label{sec:mdmarl}

	In this section, we first state formally the problem of dynamic en-route routing in the multi-agent setting.  
	Then we explain how one's sequential decision-making process can be formulated as a single-agent RL. 
	Building on this, we generalize it to multi-agent RL and then propose an efficient algorithm. 
	%we introduce a flow-dependent  mean field multi-agent deep Q-learning approach to tackle the multi-driver route choice task. 

\begin{defn}
    (MRG Problem Statement): There are $N$ intelligent agents (autonomous vehicles (AV) for example) indexed by $i\in\{1,2,\cdots,N\}$ moving from a set of origins, denoted as $\cal O$, to a set of destinations $\cal D$. 
    Each agent aims to select its optimal next-go-to edge at every intermediate node by minimizing a travel cost functional over a predefined planning horizon $[0,T]$. 
    The traffic system evolves according to all agents' actions and some transition probability at each time step. With the system state updated, those agents at intermediate nodes select the next-go-to edge. The process continues till either all agents reach the destination or the planning time horizon is terminated.   
\end{defn}
When one solves its own optimal control problem while others are doing so simultaneously, a congestion (routing) game forms. 
We assume that the system evolution dynamic remains unknown to these agents. Thus they have to learn the game and select their individual routing strategy online via their interaction with other agents and the traffic environment. The outcome is a Nash equilibrium that represents a scalable decentralized online routing strategy for each controllable agent.

% \tcr{In the multi-agent RL setting, 
% agent $i \in \{1,2,\cdots,N\}$ uses a policy $\pi_i:O_i \times A_i \rightarrow [0,1]$ to choose actions after drawing observation $o_i$. 
% After all agents taking actions, the joint action $\mathbf{a}$ triggers a state transition $\mathbf{s} \rightarrow \mathbf{s}'$ based on the state transition probability $P(\mathbf{s}'|\mathbf{s}, \mathbf{a})$. Agent $i$ draws a private observation $o_i'$ corresponding to $\mathbf{s}'$ and receives a reward $r_i(\mathbf{s}, \mathbf{a}, \mathbf{s}')$. 
% Agent $i$ aims to maximize its discounted expected cumulative reward by deriving an optimal policy $\mu^*_i$. 
% This process repeats until agents reach their own terminal state.}

\begin{rem}
\begin{enumerate}
    \item In this paper, we assume travelers are perfectly rational agents in the context of selfish routing. In other words, every traveler is a self-interest agent who aims to maximize individual accumulative rewards or minimize individual cost or maximize payoffs, including but not limited to autonomous vehicles, connected and automated vehicles, or human-driven vehicles with intelligent navigation systems and human drivers follow the navigation instructions completely. The device or mechanism that executes routing algorithms can be smart computers residing in autonomous vehicles or connected and automated vehicles, intelligent navigation aids or phone apps, and even human drivers who are perfectly rational in dynamic routing. 
    \item We use agents, players, travelers, and controllable vehicles interchangeably, which all represent those intelligent vehicles that select routes based on utility-optimizing behavior. 
\end{enumerate}
\end{rem}

\subsection{Single-agent reinforcement learning}  \label{sec:single}

	As a stepping stone, we first introduce a single-agent RL approach. Within the single-agent scope, there is only one agent interacting with the stochastic environment. The goal of the deep Q-learning approach is to derive an optimal policy so that the agent could get the maximum expected cumulative reward by following the policy in the environment. 
	
\begin{figure}[H]
		\centering
		\includegraphics[width=0.99\linewidth,height=0.25\textheight,keepaspectratio]{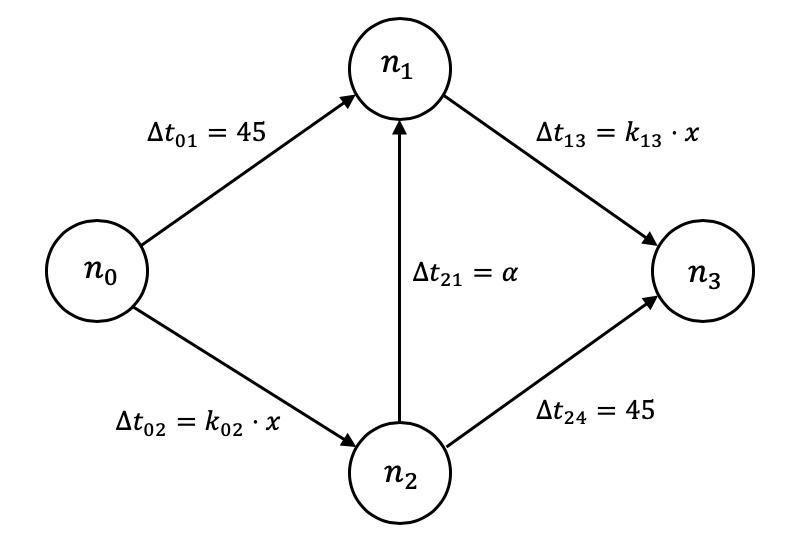}
		\centering 
		\caption{Braess network}
		\label{fig:b1graph}
	\end{figure}

	To be specific to the route choice task, we introduce the single-agent deep Q learning approach on a Braess network, as presented in Figure~(\ref{fig:b1graph}). In the Braess network, there are four nodes, namely $\{n_0,n_1,n_2,n_3\}$. $n_0$ is the origin node and $n_3$ the destination/terminal node.  There are five directed links connecting these nodes. We denote the link connecting $n_i$ and $n_j$ as $l_{ij}$. Link travel time on $l_{ij}$ is denoted by $\Delta t_{ij}$. In this study, $\Delta t_{01} = 45$, $\Delta_{23} = 45$, $\Delta t_{02} = k_{02}\cdot x$, $\Delta t_{13} = k_{13} \cdot x$, and $\Delta t_{21} = \alpha$, where $x$ denotes the travel flow (i.e., number of vehicles) on the link, $k_{02}$ and $k_{13}$ are two multipliers, and $\alpha$ is a control parameter. 
	
	 %We now illustrate some notations to be used in the single-agent RL approach. 
	 First and foremost, the interaction between the agent and the environment is typically characterized by a Markov decision process or an MDP for short \citep{puterman_markov_1994}. An MDP is specified by a tuple $(S,A,R,P,\gamma)$, where $S$ denotes the state space, $A$ the action space, $R$ the reward function, $P$ the state transition probability matrix, and $\gamma \in [0,1]$ the discount factor. An MDP goes as follows. At any state $s \in S$ but some terminal state, the agent chooses action $a \in A$ and executes the action in the environment. It observes a state transition from $s$ to a new state $s' \in S$ with probability $P(s'|s,a)$ and receives a reward $r(s,a,s') \in R$. In the context of route choice, 
	
	\begin{itemize}
		
		\item $s\in S$. State $s$ consists of two components, namely node and time, i.e., $s = (n, t)$. %For example, the initial state of the agent is $s_0 = (n_0, 0)$.
	
		\item $a \in A$.  Action $a$ for the agent currently in state $s = (n,t)$ is simply one of the outbound links from node $n$. 
		For example, $a = l_{n,n'}$, where $l_{n,n'}$ is the link connecting node $n$ and node $n'$. %For example, allowable actions for the agent at initial state $s_0$ are $l_{01}$ and $l_{02}$.
		%Equivalently, we use the end node of outbound links as allowable actions in this study. 
		In this study, we assume action is deterministic, meaning that the agent will enter the chosen link. 
		Note that the route choice for each agent is a discrete action space and the number of outbound links from a node varies from node to node in the network, indicating that the number of allowable actions is dependent on the node where the agent is currently located. 
		%For example, there are two allowable actions when the agent is at $n_0$ or $n_2$ and one allowable action when the agent is at $n_1$.
		%\item 
	
		\item $P$. After taking action $a$ in state $s$, the agent arrives at a new state $s' = (n',t')$ with probability $P(s'|s,a)$, where $n'$ is the end node of the chosen link (i.e., $l_{n,n'}$), and $t'$ is the time right after the state transition. $P$ is typically unknown to the agent. Therefore, the agent needs to repeatedly interact with the environment to gain state transition experiences, i.e., $(s,a,s')$. %For example, if the agent chooses action $l_{02}$ in initial state $s_0$, the agent arrives at $s'=(n_2,1)$ with probability $P(s'=(n_2,1)|s_0,a=l_{02}) = 100\%$. The time component in $s'$ is $1$ because the link travel time $\Delta t_{02} = x = 1$ (i.e., there is one agent on this link). The state transition probability is $100\%$ because both the action and the state transition are deterministic in this case.  
		%\item 
		
		\item $r \in R$. In addition to the observed state transition $s \rightarrow s'$, the agent receives a reward $r_{t'}(s,a,s')$ after executing action $a$, where the subscript $t'$ explicitly denotes that the reward is received by the agent at time $t'$.
		In the context of route choice, the reward function $r_{t'}(s,a,s')$ is typically chosen as some negative travel cost related to the  state transition $s \rightarrow s'$. For example, $r_{t'}(s,a,s') = -(t'-t)$, i.e., negative travel time, and $r_{t'}(s,a,s') = -\text{dist}(n,n')$, where $\text{dist}(n,n')$ measures the distance between node $n$ and $n'$, i.e., negative travel distance. 
		%\item $r\in R$. After taking action $a$ in state $s$, the agent arrives at a new state $s'$ and receives a reward $r$. We simply take the negative link travel time as the reward throughout this study. For example, if the agent chooses action $l_{02}$ in initial state $s_0$, the agent receives a reward of $-\Delta t_{02} = -1$. The goal of the agent is to maximize her expected cumulative reward, meaning that the agent aims to minimize cumulative travel time. 
		%\item 
	
		\item $\gamma$. The discount factor $\gamma$ is used to discount the future reward. When $\gamma = 1$, the agent does not differentiate future rewards from immediate rewards. As $\gamma$ get smaller, the agent cares less about rewards received in the distant future, therefore her decision-making gets more myopic.  In this study, we take $\gamma = 1$ because usually drivers aim to minimize their cumulative travel cost for a trip. In other words, they value a future travel cost and an immediate cost similarly. 
		
		\item $\rho = \sum_{t = 0}^T \gamma^t r_t$. $\rho$ is the discounted cumulative reward. The agent aims to maximize $\rho$ by deriving an optimal policy. 
	\end{itemize}

	% characterizes the stochastic environment and is typically
	%A Markov Decision Process or an MDP for short \citep{puterman_markov_1994} underlies reinforcement learning problems. 
	 %When the MDP is known, various algorithms such as dynamic programming and value iteration can be adopted to solve for the optimal policy. When the MDP is unknown, the agent repeatedly interacts with the stochastic environment and learns towards the optimal policy. This learning process goes as follows. At any state $s \in S$, the agent chooses action $a \in A$ and executes the action in the environment. It observes a state transition from $s$ to a new state $s' \in S$ with probability $P(s'|s,a)$ and receives a reward $r(s,a,s') \in R$. It collects the experience tuple $(s,a,s',r)$. This process repeats until the agent arrives at some terminal state. 

	%The agent uses the collected experience tuples, i.e., $(s,a,s',r)$, to derive an optimal policy. 
	A \emph{policy} $\mu$ is a mapping from state $s$ to action $a$, i.e., $\mu(s) = a$. %Starting from state $s$ and following policy $\mu$, the agent observes $K$ state transitions, namely $\{\mathcal{T}_1, \mathcal{T}_2, \cdots, \mathcal{T}_K\}$ before reaching some terminal state, and receives rewards $\{r_1, r_2, \cdots, r_K\}$.  Assuming reward $r_k$ is received by the agent at time $t_k$ for $k \in \{1,2,\cdots,K\}$ and agent is at time $t_s$ in state $s$, the value function of state $s$ under policy $\mu$, denoted as $V^{\mu}(s)$, is thereafter defined as the expected discounted cumulative reward, i.e., $V^{\mu}(s) = \mathcal{E}_{}\sum_{k=1}^{K} \gamma ^{t_k - t_s} r_k$. The optimal policy, denoted as $\mu^*$, can then be derived as $\mu^* = argmax_{\mu} V^{\mu}(s),~\forall s \in S$. Therefore the optimal value function $V(s) = V^{\mu^*}(s)$. 
	Following policy $\mu$, value function $V^{\mu}(s)$ is defined as the expected discounted cumulative reward from state $s$. Mathematically, it is given in a recursive form, namely $V^{\mu}(s) = \mathbb{E}_{a\sim \mu(s), s'\sim P(\cdot|s,a)}[r(s,a,s')+\gamma V^{\mu}(s')]$. 
	
	An \emph{optimal policy}, denoted as $\mu^*$, is the policy that maximizes the value function, i.e., $\mu^* = argmax_{\mu} V^{\mu}(s),~\forall s \in S$. The optimal value function, denoted as $V(s)$, is given as \citep{sutton_introduction_1998}

	\begin{equation}
	V(s) = max_a \mathbb{E}_{s'\sim P(\cdot|s,a)}[r(s,a,s')+\gamma V(s')].
	\label{eqn:v}
	\end{equation}

	While the value function $V(s)$ captures the optimal expected cumulative reward that an agent can earn from state $s$,  the state-action value $Q(s,a)$ forces the agent to take action $a$ and therefore captures the optimal expected cumulative reward from state $s$ and action $a$. 
	Mathematically, $V(s) = max_{a} Q(s,a)$. Substituting the relation between $V$ and $Q$ into Equation~(\ref{eqn:v}) yields	

	\begin{equation}
	Q(s,a) =\mathbb{E}_{s'\sim P(\cdot|s,a)}[r(s,a,s')+\gamma max_{a'}Q(s',a')].
	\label{eqn:q}
	\end{equation}
	With the optimal action-value function $Q$, optimal policy can be explicitly derived as $\mu^*(s) = argmax_a Q(s,a)$. 

	With the collected experience tuple $(s,a,s',r)$, the agent can update the $Q$ value by the widely used tabular Q-learning algorithm 
	\begin{equation}
	Q(s,a) \leftarrow Q(s,a) + \eta [r + \gamma max_{a'}Q(s',a') - Q(s,a)],
	\label{eqn:tabularq}
	\end{equation}
	where $\eta \in (0,1]$ is the learning rate. Convergence is guaranteed for the Q-learning algorithm when the learning rate is decayed properly over time \citep{sutton_introduction_1998}. 
	%\tcb{Unfortunately, the tabular Q-learning algorithm in Equation~(\ref{eqn:tabularq}) is only applicable to small and discrete state and action spaces. It is not feasible to be used in a problem with large state and action spaces.} 
	%Tabular Q-learning maintains a Q-table for all possible combinations of state and action. 
	%It is thus not feasible to be used in a problem with \tcb{a large or continuous state space and a large action space}. 
	Thanks to the emergence of deep Q networks (DQNs) \citep{mnih_human-level_2015}, we could resort to neural networks as a functional approximator of the $Q$ function. Denoting the functional approximator parameterized by $\theta$ as $Q_{\theta}$, DQN updates its parameter $\theta$ by minimizing the following loss
	\begin{equation}
	\mathcal{L}(\theta) = \mathbb{E}_{s,a,s'}\left[ \left(r(s,a,s')+\gamma max_{a'} Q_{\theta^-}(s',a') - Q_{\theta}(s,a) \right)^2 \right],
	\label{eqn:qloss}
	\end{equation}
	where $Q_{\theta^-}$ is called a target network copied from $Q_{\theta}$ every $\tau$ episodes to ensure training stability, where $\tau$ is a hyperparameter.

	\begin{exmp} (Single-agent route choice).
		%To be more concrete, we demonstrate the aforementioned notations by a Braess network shown in Figure~\ref{fig:b1graph}. Note that we use the Braess network throughout this study and will revisit Figure~\ref{fig:b1graph} in later sections. 
		
		%\begin{itemize}
		%\item 
		To be more concrete, we demonstrate the aforementioned notations in the Braess network. Supposing an agent is initially placed at $n_0$ in the Braess network, the initial state of the agent is $s = (n_0, 0)$. There are two allowable actions for the agent, namely $l_{01}$ and $l_{02}$. Assuming the agent chooses action $l_{02}$ in the initial state, the agent arrives at $s'=(n_2,1)$ with probability $P(s'=(n_2,1)|s=(n_0,0),a=l_{02}) = 100\%$. The time component in $s'$ is $1$ because the link travel time $\Delta t_{02} = x = 1$ (i.e., there is one agent on this link). The state transition probability is $100\%$ because both the action and the state transition are deterministic in this case. Assuming the reward function is the negative travel time, the agent receives a reward of $-\Delta t_{02} = -1$ along with the state transition $s\rightarrow s'$.
		
		We now apply the single-agent Q-learning to the Braess network with one agent initially at $n_0$. For the purpose of demonstration, we assume $\alpha=1$ and $k_{02} = k_{13} = 40$ in this example. %The agent repeatedly interacts with the environment and collects experience tuples. Using Equation~(\ref{eqn:qloss}), the agent updates $\theta$ until convergence. 
		Due to the simplicity of this case, it could be solved analytically from the destination and moving backwards. Noticing the static nature of this example, we neglect the time component in state. 
		$n_3$ is the terminal node, meaning that $V(n_3) = 0$. 
		There are two nodes connecting $n_3$, namely $n_1$ and $n_2$. Note that the agent could arrive at $n_1$ from $n_2$, meaning that $V(n_2)$ is dependent on $V(n_1)$ and $V(n_1)$ is independent on $V(n_2)$, we thus discuss $n_1$ first. At node $n_1$, the only allowable action is $l_{13}$ leading the agent to terminal node $n_3$, yielding $Q(n_1,l_{13}) = -\Delta t_{13} + \gamma V(n_3) = -40$ and $V(n_1)=-40$. At node $n_2$, there are two actions, namely $l_{21}$ and $l_{24}$. Action $l_{21}$ leads the agent to $n_1$, yielding $Q(n_2,l_{21}) = -\Delta t_{21} + \gamma V(n_1) = -(\alpha + 40) = -41$. Action $l_{24}$ leads the agent to the terminal node $n_3$, yielding $Q(n_2,l_{24}) = -\Delta t_{24} + \gamma V(n_3) = -45$. Therefore, $V(n_2) = max(Q(n_2,l_{21}) , Q(n_2,l_{24}) ) = -41$ and $\mu^*(n_2) = l_{21}$ . At node $n_0$, there are two actions, namely $l_{01}$ and $l_{02}$. Action $l_{01}$ leads the agent to $n_1$, yielding $Q(n_0,l_{01}) = -\Delta t_{01} + \gamma V(n_1) = -85$. Similarly, $Q(n_0,l_{02}) = -\Delta t_{02} + \gamma V(n_2) = -81$. Therefore,  $V(n_0) = max(Q(n_0,l_{01}), Q(n_0,l_{02})) = -81$ and $\mu^*(n_0) = l_{02}$. Following optimal policy $\mu^*$, the agent takes route $n_0 \rightarrow n_2 \rightarrow n_1 \rightarrow n_3$, which is the shortest path in this example. 
		%Based on these derived Q values at every state, the optimal policy $\mu^*(s) = argmax_a Q(s,a)$, $\forall s \in S$.  
		%Q values of interest are listed in Table~\ref{tab:qvals}.
	\end{exmp}

\subsection{Multi-agent reinforcement learning and Markov games} \label{sec:multi}

	Although the single-agent Q learning efficiently finds the shortest path, it fails to capture the competition among agents in a multi-agent system (MAS) \citep{di2021survey}. In a single-agent RL problem, only one agent is placed in the environment to learn the optimal policy, which maximizes the expected cumulative reward of the agent. Unfortunately, when multiple agents follow this optimal policy, their expected cumulative reward might be low. For example, if there are 50 agents at node $n_0$ initially in the Braess network, their expected cumulative reward is $-110$ by following the optimal policy (i.e., the shortest path $n_0 \rightarrow n_2 \rightarrow n_1 \rightarrow n_3$). This reward (i.e., $-110$) is lower than that of following another route (e.g., $n_0 \rightarrow n_1 \rightarrow n_3$ yields an expected cumulative reward of $-95$).  Therefore, to tackle a multi-driver route choice task, we develop a multi-agent deep Q-learning approach in this section to capture the competition among agents.

\subsubsection{Problem formulation}

%\textcolor{red}{remember to define Markov Game, echoing Sec }

	Similar to the previous single-agent case, we introduce how to formulate the multi-agent routing problem on the Braess network shown in Figure~(\ref{fig:b1graph}). %Without loss of generality, two agents are initially placed at node $n_0$. 

	Due to the existence of multiple agents, the multi-agent routing problem is formulated as a Markov game \citep{littman_markov_1994}, %partially observable Markov decision process (POMDP) \citep{littman_markov_1994}, 
	which is a generalization of Markov decision processes to multiple interacting agents with competing goals, in which the environment makes transitions probabilistically in response to the agents' actions.
%	We denote the Markov game by a tuple $(N,S,O,A,P,R,\gamma)$,  where $N,S, O, A, P, R, \gamma$ are the number of agents, environmental state space, joint private observation space, joint action space, state transition probability functions, reward functions, and the discount factor, respectively. 
We denote the Markov game by a partially observable Markov decision process \citep{littman_markov_1994}, defined by a tuple $(S, O_1, O_2, \cdots, O_N, A_1, A_2, \cdots, A_N, P, R_1, R_2, \cdots, R_N, N, \gamma)$, where $N$ is the number of agents and $S$ is the environment state space. Environment state $\mathbf{s} \in S$ is not fully observable. Instead, agent $i$ draws a private observation $o_i \in O_i$ which is correlated with $\mathbf{s}$. $O_i$ is the observation space of agent $i$, yielding a joint observation space $O = O_1 \times O_2, \times \cdots \times O_N$, $A_i$ is the action space of agent $i \in \{1,2,\cdots, N\}$, yielding a joint action space $A = A_1 \times A_2 \times \cdots \times A_N$, $P:S\times A \times S \rightarrow [0, 1]$ is the state transition probability, $R_i:S\times A \times S \rightarrow \mathbb{R}$ is the reward function for agent $i$, and $\gamma$ is the discount factor. 
Agent $i \in \{1,2,\cdots,N\}$ uses a policy $\pi_i:O_i \times A_i \rightarrow [0,1]$ to choose actions after drawing observation $o_i$. %Policies of all agents form a joint policy $\boldsymbol{\pi}$. 
After all agents taking actions, the joint action $\mathbf{a}$ triggers a state transition $\mathbf{s} \rightarrow \mathbf{s}'$ based on the state transition probability $P(\mathbf{s}'|\mathbf{s}, \mathbf{a})$. Agent $i$ draws a private observation $o_i'$ corresponding to $\mathbf{s}'$ and receives a reward $r_i(\mathbf{s}, \mathbf{a}, \mathbf{s}')$. 
Agent $i$ aims to maximize its discounted expected cumulative reward by deriving an optimal policy $\mu^*_i$. %which is the best response to other agents' policies.
This process repeats until agents reach their own terminal state.

Due to the existence of other agents, the Q-value function for agent $i$ %following the policy $\pi_i$
, i.e., $Q_i$, is now dependent on the environment state $\mathbf{s} \in S$ and the joint action $\mathbf{a} \in A$ of all agents%, and the joint policy $\boldsymbol{\pi}$
, i.e, 
\begin{equation}
Q_i = Q_i(\mathbf{s}, \mathbf{a}). \label{eqn:q1}
\end{equation}
Similarly, the value function of agent $i$, i.e., $V_i = V_i(\mathbf{s})$, %following policy $\pi_i$ 
is dependent on the environment state $\mathbf{s}$. 

In the multi-agent system, \emph{optimal policy} of agent $i$, denoted as $\mu_i^*$, is defined as the $i^{th}$ agent's best response to others' policies when others hold their policies constant. Below we introduce the formal definition of optimal policy in a multi-agent system using the concept of Nash equilibrium \citep{littman2001value}.

\begin{defn}
\begin{enumerate}
    \item 
     A set of policies $\mu_i, \cdots, \mu_N$ is a Nash equilibrium if each is a best response to the others. In other words, nobody can improve her value function by unilaterally switching her policy, while all others hold their policies fixed.
     That is, $\forall i\in \{1,\cdots, N\}$, the value attained by agent $i$ from any state $s$,
     %The optimal value function, denoted as $V(s)$, is given as \citep{sutton_introduction_1998}
	\begin{equation}
	V_i(s) = max_{a_i} 
	\mathbb{E}_{\mathbf{a}^{-i}\sim A}
	\mathbb{E}_{s'\sim P(\cdot|s,\mathbf{a})}[r_i(s,\mathbf{a},s')+\gamma V_i(s')].
	\label{eqn:v1}
	\end{equation}
	At a Nash equilibrium, each agent maximizes its value function given that all other agents remain fixed. 
	Every Markov game has a Nash equilibrium in stationary policies \citep{filar1997applications}.
	\item \emph{Markov strategy} is the policy that depends only on the current state. 
	If all players other than $i$ play Markov strategies, then player $i$ has a best response that is a Markov strategy \citep{solan2015stochastic}. %johari_2007,
	In this paper, we mainly focus on Markov strategy that is a mapping from the current state $s$ to action $\mathbf{a}$.
\end{enumerate}
\end{defn}
%The policy that maximizes the value function, i.e., $\mu_i^* = argmax_{\mu} V_i^{\mu}(s),~\forall s \in S$. 

	Multi-agent RL is a computational tool for Markov games. We further specify each component using the language of MARL below. 
	\begin{itemize}
		\item $N$. There are $N$ controllable adaptive agents, denoted by $\{1,2,\cdots,N\}$. Note that in a traffic network, there are also other traffic participants who are not controlled by the learning algorithm developed in this study. We regard those uncontrollable traffic participants as background traffic. The background traffic affects the behavior of controllable agents by its impact on link travel cost. For example, if the background traffic has caused a traffic jam on a link, it is very likely that controllable agents will avoid this jammed link. 
		\item $\mathbf{s} \in S$. Environmental state $\mathbf{s}$ consists of some global information such as distribution of agents and traffic condition on each link. $\mathbf{s}$ is not fully accessible to agents. %We denote the initial environmental state by $\mathbf{s}_0$.
		\item $\mathbf{o} \in O$. Although the environmental state $\mathbf{s}$ is not observable by agent $i$, the agent is able to draw a private observation $o_i \in O_i$, which is correlated with $\mathbf{s}$. The Cartesian product of private observation spaces of all agents forms the joint observation space, i.e., $O = O_1 \times O_2 \times \cdots \times O_N$. In this paper, $o_i$ consists of two components, namely node and time, i.e., $o_i = (n,t)$. Joint observation $\mathbf{o} = (o_1,o_2, \cdots, o_N)$. %For example, at current time $t = 0$, $\mathbf{o} = ((n_0,0),(n_0,0))$. 
		\item $\mathbf{a} \in A$. Based on $o_i = (n,t)$, the allowable action set for agent $i \in \{1,2,\cdots,N\}$ consists of all outbound links from node $n$. For example, $a_i = l_{n,n'}$, where $l_{n,n'}$ is the link connecting $n$ and $n'$. Joint action $\mathbf{a} = (a_1,a_2,\cdots,a_N)$. %at current time $t=0$, the allowable action set for both agents is $\{l_{01}, l_{02}\}$. Denoting actions taken by agents are $a_1$ and $a_2$, respectively, joint action $\mathbf{a} = (a_1,a_2)$. The Cartesian product of action spaces of all agents forms the joint action space, i.e., $A = A_1 \times A_2 \times \cdots \times A_N$. 
		
		\item $P$.  Joint action $\mathbf{a}$ triggers a state transition $\mathbf{s} \rightarrow \mathbf{s}'$ with probability $P(\mathbf{s}'|\mathbf{s}, \mathbf{a})$. Agent $i \in \{1,2,,\cdots,N\}$ draws a new private observations, namely $o_i' = (n',t')$, where $n'$ is the end node of the chosen link $l_{n,n'}$ and $t'$ is the time when agent $i$ arrives at $n'$. The private observation $o_i'$ is correlated with $\mathbf{s}'$. 
		
		%Agents draw new private observations, namely $o_1'$ and $o_2'$, respectively, based on $\mathbf{s}'$. For example, if joint action $\mathbf{a} = (l_{01}, l_{02})$, then $o_1' = (n_1, \Delta t_{01} = 45)$ and $o_2' = (n_2, \Delta t_{02} = k_{02})$; If joint action $\mathbf{a} = (l_{02}, l_{02})$, then $o_1' = (n_2, \Delta t_{02} = 2\cdot k_{02})$ and $o_2' = (n_2, \Delta t_{02} = 2 \cdot k_{02})$. Note that although agent $2$ takes action $l_{02}$ in both cases, she arrives at different observations, i.e., $(n_2, k_{02})$ and $(n_2, 2\cdot k_{02})$ due to the changing behavior of agent $1$ (i.e., taking action $l_{01}$ in the first case and $l_{02}$ in the second). 
		
		\item $r\in R$. In addition to $o_i'$, agent $i \in \{1,2,\cdots,N\}$ receives a reward $r_{i,t'}(\textbf{s},\textbf{a}, \textbf{s}')$, where the subscript $i,t'$ explicitly denotes that the reward is received by agent $i$ at time $t'$. 
		 Similar to the single-agent RL, reward is typically some negative travel cost such as travel time and travel distance in the context of route choice.  %For example, if joint action $\mathbf{a} = (l_{01}, l_{02})$, then $r_1 = -\Delta t_{01} = -45$ and $r_2 = - \Delta t_{02} = - k_{02}$; If joint action $\mathbf{a} = (l_{02}, l_{02})$, then $r_1 = - \Delta t_{02} = -2\cdot k_{02}$ and $r_2 = - \Delta t_{02} = - 2 \cdot k_{02}$.
		\item $\gamma$. Similar to the single-agent RL, $\gamma = 1$.
		\item $\rho_i = \sum_{t=0}^T \gamma^t r_{i,t}$. $\rho_i$ is the discounted cumulative reward received by agent $i$. If the goal of agent $i$ is to maximize $\rho_i$, agents are non-cooperative and selfish; if the goal of agent $i$ is to maximize the average discounted cumulative rewards of all agents, i.e., $\frac{1}{N} \sum_{i=1}^N \rho_i$, agents are cooperative. %by deriving optimal policy $\mu^*_i$. %Therefore, agents are non-cooperative and selfish in this study. In a cooperative environment, agents usually aim to maximize the overall expected cumulative reward of all agents. 
	\end{itemize}

%Joint action of all agents, i.e., $\textbf{a} = (a_1,a_2,\cdots,a_N)$ triggers a state transition $\textbf{s} \rightarrow \textbf{s}'$ with probability $P(\textbf{s}'|\textbf{s}, \textbf{a})$ in the environment. Agent $i \in \{1,2,\cdots,N\}$ receives a reward $r_i(\textbf{s},\textbf{a}, \textbf{s}')$ and draws a new private observation $o_i'$ from environmental state $\textbf{s}'$. 
%The decision-making process (i.e., every agent chooses an action whenever draws a new private observation) repeats until reaching some terminal state. Similar to the single-agent case, the goal of agent $i$ is to maximize her expected cumulative reward by deriving optimal policy $\mu^*_i$. Therefore agents are non-cooperative and selfish in this study. In a cooperative environment, agents usually aim to maximize the overall expected cumulative reward of all agents. 

	Due to the coexistence of other agents, $Q$ function of agent $i$, i.e., $Q_i$ is now a function of environmental state $\mathbf{s}$ and joint action $\mathbf{a}$, namely
	\begin{equation}
	Q_i = Q_i(\mathbf{s}, \mathbf{a}).
	\label{eqn:qi}
	\end{equation}
	Similarly, value function of agent $i$, i.e., $V_i$ is a function of environmental state $\mathbf{s}$, namely $V_i = V_i(\mathbf{s})$. Note that each agent $i$ has her own $Q$ function, i.e., $Q_i$, which may be different from $Q$ functions of other agents, and therefore has an optimal policy $\mu^*_i$ different from policies of other agents. This distinguishes multi-agent RL from its single-agent counterpart, because agents may behave differently even in the same state with multi-agent RL while they choose the same action in the same state with single-agent RL. 
	
	Considering that environmental state $\textbf{s}$ is not fully observable by agents, Q function shown in Equation~(\ref{eqn:qi}) is not tractable from the perspective of agent $i$. Actually, it is private observation $o_i$ based on which agent $i$ chooses next action. 
	We thus rewrite Q function for agent $i$ as
	\begin{equation}
	Q_i = Q_i(o_i, o_{-i}, a_i, a_{-i}),
	\label{eqn:qi_}
	\end{equation}
	where $o_{-i}$ and $a_{-i}$ denote the joint observation and joint action of all agents except agent $i$, respectively. In a non-coordinated environment (i.e., agents are not sharing their private observations or actions), $o_i$ and $a_i$ are but $o_{-i}$ and $a_{-i}$ are not accessible to agent $i$. Unfortunately, removing the dependency of $Q_i$ on $a_{-i}$ or $o_{-i}$ may introduce large instability in Q-learning. In addition, convergence guarantee in single-agent Q-learning is no longer valid due to the adaptive behavior of other agents if $o_{-i}$ and $a_{-i}$ are not included \citep{matignon_independent_2012}.

    For a real-world multi-agent problem, there are commonly more than tens of thousands of agents. Maintaining one Q function, i.e., $Q_i$ for each agent is thus computationally infeasible. Leveraging some homogeneity among agents (e.g., some agents share the same utility function), we broadly categorize agents into $C$ groups and enable Q function sharing within each group. We denote the shared Q function for agents in group $c \in \{1,2,\cdots,C \}$ as
    \begin{equation}
 	Q^c(o_i,o_{-i},a_i,a_{-i}). 
 	\label{eqn:qc}
 	\end{equation}

	The dimension of the input to the shared Q function easily becomes prohibitively large when the number of agents in the environment increases. 
	For example, the joint action space for 1,000 agents is $1,000 \times d$ dimensional if each action is a $d$ dimensional vector. We take $d$ as a generic representation of the dimension of the action space of each agent. For example, d could be 1 action of agents denoted as a scalar (e.g., 1,2,..., i.e., label encoding) or 3 when the agent has three actions to take and these actions are represented by one hot encodings. To tackle this issue, we first notice that joint observation $o_{-i}$ and joint action $a_{-i}$ record full information of other agents, which may contain redundant information from the perspective of agent $i$. Actually, in the task of route choice, observations and actions of agents who are currently far away from agent $i$ have a very limited influence on agent $i$. It is nearby agents who exert an impact on the traffic condition around agent $i$. For example, after agent $i$ chooses a link at a node, an immediate influential quantity is the traffic flow on the chosen link. If the flow is high, agent $i$ may experience a high travel cost on the link; if the flow is low, agent $i$ may have a smooth transition to the end of the link.  
	Therefore, high-dimensional $o_{-i}$ and $a_{-i}$ in Equation~(\ref{eqn:qc}) could be approximated by some low-dimensional aggregate information of nearby agents. Similar to the mean field approximation in \cite{yang_mean_2018}, we call this aggregate information as the mean action of nearby agents and denote it by $\bar{a}_i$. $Q$ function in Equation~(\ref{eqn:qc}) thus becomes
	\begin{equation}
	Q^c(o_i,o_{-i},a_i,a_{-i}) \approx Q^c(o_i,a_i,\bar{a}_i).
	\label{eqn:mfq}
	\end{equation}

	%To address this issue, we adopt the mean field approximation \citep{yang_mean_2018} to simplify the interaction among agents. Noticing that observations and actions of agents who are currently away from agent $i$ have a very limited impact on the $Q$ function of agent $i$,  we therefore use a mean action, denoted as $\bar{a}_i$, to serve as a proxy of $a_{-i}$. The mean action $\bar{a}_i$ essentially captures the interaction between agent $i$ and her neighbor agents. With the mean action $\bar{a}_i$, $Q$ function of agent $i$ in group $c$ becomes

	We now formally define the mean action in the context of route choice. 
	\begin{defn} (\emph{Mean action}). 
		Mean action $\bar{a}_i$ is defined as the traffic flow on the link that is chosen by agent $i$.
		Note that the traffic flow is calculated right after agent $i$ enters the link.
		%For example, in the Braess network with two agents initially at node $n_0$, if joint action $\mathbf{a} = (l_{01}, l_{02})$, then $\bar{a}_1 = 1$ and $\bar{a}_2 = 1$; If joint action $\mathbf{a} = (l_{02}, l_{02})$, then $\bar{a}_1 = \bar{a}_2 = 2$. 
	\end{defn}

% 	With the above definition of the mean action, we interpret the right hand side of Equation~(\ref{eqn:mfq}), i.e., $Q^c(o_i,a_i,\bar{a}_i)$, as the flow-dependent $Q$ function. 
	
% 	\begin{defn} (\emph{Flow-dependent $Q$ function}). 
% 		 With agent $i$ choosing action $a_i$ at observation $o_i$, the explicit inclusion of mean action $\bar{a}_i$, i.e., the traffic flow on the chosen link, makes $Q^c(o_i,a_i,\bar{a}_i)$ a flow-dependent $Q$ function. 
% 	\end{defn}

    % \tcb{moved to detailed explanation of shared Q function to appendix. we may consider deleting this part if not necessary.  }

%\subsubsection{Optimal route choice}

%	With the flow-dependent $Q$ function, the optimal policy of agent $i$ in group $c \in \{1,2,\cdots,C\}$ is 
%	\begin{equation}
%		\mu^*_i (o_i) = argmax_{a_i} \mathbb{E}_{\bar{a}_i \sim \mu^*_{-i}}[Q^c(o_i, a_i, \bar{a}_i)],~\forall o_i \in O_i.
%		\label{eqn:poli}
%	\end{equation}

%	\textcolor{red}{flow-dependent, aBar is the flow}
	
%	\textcolor{red}{It is the inclusion of aBar making the shared Q feasible, agents could have different actions}
	
%	\textcolor{red}{compared to actor-critic, this approach could automatically address the problem of varibale action set}

\subsubsection{Mean field multi-agent deep Q-learning}

%\textcolor{blue}{Flow-dependent multi-agent deep Q learning? using MF-MA-DQL?}

	Different from single-agent RL, agent $i$ in group $c \in \{1,2,\cdots,C\}$ in an MAS interacts with not only the environment but also other agents and collects experience tuples in the form of $(o_i,a_i,o_i',r_i,\bar{a}_i)$. Considering that mean action $\bar{a}_i$ is typically in a large discrete or even continuous space, a DQN parameterized by $\theta_c$, denoted by $Q^c(o_i,a_i,\bar{a}_i|\theta_c)$, is used to approximate $Q^c(o_i,a_i,\bar{a}_i)$.  Similar to Equation~(\ref{eqn:qloss}), DQN updates its parameter $\theta_c$ by minimizing the following loss
	\begin{equation}
	\mathcal{L}(\theta_c) = \mathbb{E}_{o_i,a_i,\bar{a}_i,o_i'}\left[ (r_i + \gamma max_{a_i'} \mathbb{E}_{\bar{a}_i' \sim \mu^*_{-i}}[Q^c(o_i',a_i',\bar{a}_i'|\theta^-_c)] - Q^c(o_i,a_i,\bar{a}_i|\theta_c))^2\right]
	\label{eqn:cqloss}
	\end{equation}
	where $Q^c(\cdot|\theta^-_c)$ is a target network copied from $Q^c(\cdot|\theta_c)$ every $\tau$ episodes to stabilize training. After updating $Q$ function, optimal policy
	\begin{equation}
	\mu_i^*(o_i) = argmax_{a_i} \mathbb{E}_{\bar{a}_i \sim \mu^*_{-i}}[Q^c(o_i, a_i, \bar{a}_i|\theta_c)],~\forall o_i \in O_i.
	\label{eqn:poli}
	\end{equation}
	
	%\emph{Remark.} Although the mean action $\bar{a}_i$ may not be fully observable by agent $i$, we include it in the flow-dependent $Q$ function, i.e., $Q^c(o_i,a_i,\bar{a}_i)$ during training. In execution, only the derived optimal policy $\mu^*_i$ which is computationally a dictionary mapping from $o_i$ to $a_i$ is used, without involving mean action $\bar{a}_i$. This is essentially the idea of the centralized training and decentralized execution paradigm where some global information is used in training but not in execution \citep{lowe_multi-agent_2017}.

	\begin{algorithm}[H]
		\caption{Mean field multi-agent deep Q-learning (MF-MA-DQL)}
		\label{alg:mfdql}
		\begin{algorithmic}[1]
			\State Input: exploration parameter $\epsilon = \epsilon_0$, learning rate $\eta = \eta_0$, target network update period $\tau$
			\State Initialize one DQN $Q^c(o,a,\bar{a}|\theta)$, parameterized by $\theta_c$, and one target network $Q^c(o,a,\bar{a}|\theta^-_c)$ for each group $c \in \{1,2,\cdots,C\}$
			\State Initialize $N$ dictionaries to store the optimal policy for agents, i.e.,  $\mu^*_i, \forall i \in \{1,2,\cdots, N\}$
			\State Initialize one experience replay buffer $B_c$ for each group $c \in \{1,2,\cdots,C\}$ % and $N$ mean action buffers, i.e., $M_i, \forall i \in \{1,2,\cdots, N\}$ 
			\State Set $episode = 0$
			\Repeat
			\State From the initial environmental state $\mathbf{s}_0$, each agent $i$ draws observation $o_i$
			\State Set $t = 0$
			%\State Initialize an episode as an empty list $e$
			\Repeat
%			\State Sample a value $x$ from a uniform distribution which is defined on $[0,1]$
%			\If{$x < \epsilon$}
%			\State Select action $a_i$ from the allowable action space randomly for all available agents
%			\Else
%			\State Select action $a_i$ greedily according to the policy $\mu^*_i$ for all available agents
%			\EndIf
			\State For each agent $i$, select action $a_i$ according to the $\epsilon$-greedy method, i.e., randomly select an allowable action with probability $\epsilon$ and greedily according to policy $\mu^*_i$ with probability $1-\epsilon$
			%\State Form the joint action $a^t = (a_1^t, a_2^t, \cdots, a_N^t)$
			\State Execute joint action $\mathbf{a} = (a_1,a_2,\cdots,a_N)$ in the environment to trigger state transition $\mathbf{s} \rightarrow \mathbf{s}'$
			\State Each agent $i$ draws new observation $o_i'$, receives reward $r_i$, and observes mean action $\bar{a}_i$
			\State Store experience tuple $(o_i,a_i,o_i',r_i,\bar{a}_i)$ into replay buffer $B_c$ if agent $i$ belongs to group $c$% and $\{(o_i,a_i):\bar{a}_i\}$ into mean action buffer $M_i$
			%\State Update $Q_i(s_i,a_i,\bar{a}_i)$ by Equation~(\ref{eqn:updateQ}) for each available agent
			%\State Add the step $(s, a, r)$ into the episode $e$
			%\State The agent transits into state $s'$, and hence set $s=s'$
			\State $t \leftarrow t + 1$
			\Until{$t = T$}
			%\State Add the episode $e$ into $dict$
			\For{$c=1~\text{to}~C$}
			\For{$j=1~\text{to}~J_c$}
			\State Sample a batch of size $K_c$ from replay buffer $B_c$
			\State Update parameter $\theta_c$ of $Q^c$ by minimizing the loss defined in Equation~(\ref{eqn:cqloss})
			\State Update optimal policy $\mu^*_i$ of agent $i$ by Equation~(\ref{eqn:poli})
			\EndFor
			\EndFor
			\State $episode = episode + 1$
			\State Decrease the exploration parameter $\epsilon$ %$\epsilon = \epsilon_0 \times e^{-\tau \times I}$
			\State Decrease the learning rate $\eta$
			\If{$episode~\text{mod}~\tau = 0$} 
			\For{$c=1~\text{to}~C$}
			\State Update parameter $\theta^-_c$ of the target network, i.e., $\theta^-_c \leftarrow \theta_c$
			\EndFor
			\EndIf
			\Until{the algorithm converges}
			\State Return optimal policies $\mu^*_i, \forall i \in \{1,2,\cdots,N\}$
		\end{algorithmic}
	\end{algorithm}

	The mean field multi-agent deep Q-learning (MF-MA-DQL) algorithm is summarized in Algorithm~(\ref{alg:mfdql}). 
	%In an MAS, state transition probability functions are typically unknown due to the adaptive behavior of other agents. We therefore resort to a model-free RL approach where agents repeatedly interact with the stochastic environment. 
	We first initialize a centralized deep $Q$ network, parameterized by $\theta_c$, and a target $Q$ network, parameterized by $\theta^-_c$ for each group $c \in \{1,2,\cdots,C\}$. Although $Q$ function is shared among agents in the same group, each agent $i$ in group $c$ has her own optimal policy $\mu^*_i$, which is a dictionary mapping from observation $o_i$ to action $a_i$. From initial environmental state $\mathbf{s}_0$, each agent $i \in \{1,2,,\cdots, N\}$ keeps drawing private observation $o_i$ and taking action $a_i$ according to the widely used $\epsilon$-greedy method (i.e., agent $i$ chooses action randomly with probability $\epsilon$ and greedily from optimal policy $\mu^*_{i}$ with probability $1 - \epsilon$), until reaching some terminal state. Joint action $\mathbf{a} = (a_1,a_2,\cdots,a_N)$ is executed in the environment to trigger the state transition $\mathbf{s} \rightarrow \mathbf{s}'$.
	%Whenever agent $i$ executes an action $a_i$ in the environment, the 
	Each agent $i$ draws new private observation $o_i'$, receives reward $r_i$, and observes mean action $\bar{a}_i$.  An experience replay buffer $B_c$ for each group $c$ is used to store experience tuple $(o_i,a_i,r_i,\bar{a}_i,o_i')$ generated by agent $i$ in group $c$. Batch training is then used to update the centralized $Q$ function by sampling a batch of size $K_c$ from buffer $B_c$ and minimizing the loss shown in Equation~(\ref{eqn:cqloss}) over this batch. With the updated centralized $Q$ function, each agent updates her optimal policy by Equation~(\ref{eqn:poli}). 

	\emph{Remark. } As a value based approach, the developed MF-MA-DQL algorithm does not suffer the problem of a variable action set. A variable action set means that the number of allowable actions might vary from state to state. For example, there are two outbound links from node $n_0$ and one outbound link from node $n_1$ in the Braess network shown in Figure~(\ref{fig:b1graph}), causing the problem of variable action set. 
	A policy based approach, e.g., the widely used actor-critic method, does not fit naturally well into a RL problem with variable action set. The main reason is that the policy neural network typically takes as input a state and outputs a probability distribution on all allowable actions (i.e., the action set). Consequently, a variable action set might result in some challenges to maintain a good structure of the policy network. 
	Fortunately, value based approaches automatically handle the problem of variable action set, because a value based approach calculates the expected $Q$ value of all possible actions at a state and selects the action with the highest $Q$ value as the optimal policy. 

	\begin{exmp} (Multi-agent route choice).
		Now we apply the MF-MA-DQL algorithm to a multi-agent route choice problem. In the Braess network shown in Figure~(\ref{fig:b1graph}), now two agents are initially placed at node $n_0$. The goal of both agents is to travel to the destination node $n_3$ as soon as possible. Consequently, these two agents are considered to be homogeneous, because they share the same reward function (i.e., the negative travel time) and go to the same destination. 
		We first demonstrate the notations of Markov game in this example. 
		
		There are $N=2$ agents in the Braess network. Environmental state $\mathbf{s}$ is not observable, 
		Correlated with $\mathbf{s}$, the joint observation of agents $\mathbf{o} = ((n_0, 0),(n_0,0))$. 
		There are two allowable actions for both agents, namely $l_{01}$ and $l_{02}$. The joint action $\mathbf{a} = (a_1,a_2)$. 
		$\mathbf{a}$ triggers state transition $\mathbf{s} \rightarrow \mathbf{s}'$ with probability $P(\mathbf{s}'|\mathbf{s}, \mathbf{a})$. Each agent $i \in \{1,2\}$ then draws new private observation $o_i'$, receives reward $r_i$, and observes mean action $\bar{a}_i$. To be precise, we illustrate $o_i'$, $r_i$, and $\bar{a}_i$ under two joint actions.
		\begin{itemize}
			\item With $\mathbf{a} = (l_{01}, l_{02})$, $o_1' = (n_1, \Delta t_{01} = 45)$ and $o_2' = (n_2, \Delta t_{02} = k_{02})$. $r_1 = -\Delta t_{01} = -45$ and $r_2 = - \Delta t_{02} = - k_{02}$. $\bar{a}_1 = 1$ and $\bar{a}_2 = 1$.
			\item With $\mathbf{a} = (l_{02}, l_{02})$, $o_1' = (n_2, \Delta t_{02} = 2\cdot k_{02})$ and $o_2' = (n_2, \Delta t_{02} = 2 \cdot k_{02})$. $r_1 = - \Delta t_{02} = -2\cdot k_{02}$ and $r_2 = - \Delta t_{02} = - 2 \cdot k_{02}$. $\bar{a}_1 = \bar{a}_2 = 2$. 
		\end{itemize}
		Note that although agent $2$ takes action $l_{02}$ in both cases, the agent arrives at different observations, i.e., $(n_2, k_{02})$ and $(n_2, 2\cdot k_{02})$ due to the changing behavior of agent $1$ (i.e., taking action $l_{01}$ in the first case and $l_{02}$ in the second). 
		
		%Different from single-agent deep Q-learning, there is mean action $\bar{a}$ in the developed multi-agent deep Q-learning approach. 
		
		Due to the simplicity of this case, we now analyze optimal value at each node in the network. For the purpose of demonstration, we assume $\alpha=10$ and $k_{02} = k_{13} = 40$ in this example.  Similar to Example 2.1, we neglect the time component in private observation due to the static nature of this problem. 
		\begin{enumerate}
			\item Noticing that node $n_3$ is the terminal node, $V(n_3) = 0$. 
			\item At node $n_1$, the only action that agents can take is $l_{13}$. After one or two agents, depending on whether both of them reach node $n_1$, taking action $l_{13}$, there are two scenarios: 1) there is only one agent (i.e., either agent $1$ or agent $2$) on link $l_{13}$, meaning that traffic flow on $l_{13}$ is $1$ and it takes the agent $\Delta t_{13} = k_{13} \cdot 1 = 40$ to reach $n_3$. Therefore, $Q(o=n_1,a=l_{13},\bar{a}=1) = -40$. Note that here we drop the subscript index $i$ for the agent for notation simplicity, because agents are homogeneous in this case and $Q(o_1=n_1,a_1=l_{13},\bar{a}_1=1) = Q(o_2=n_1,a_2=l_{13},\bar{a}_2=1) = -40$. In other words, no matter who the agent is, it takes the agent $40$ time steps to reach $n_3$ starting from $n_1$, taking action $l_{13}$, and observing $\bar{a} = 1$ ; 2) there are two agents on link $l_{13}$, meaning that traffic flow on $l_{13}$ is $2$ and it takes the agent $\Delta t_{13} = k_{13} \cdot 2 = 80$ to reach $n_3$. Therefore, $Q(o=n_1,a=l_{13}, \bar{a} = 2) = -80$. Note that in scenario 2), it could be either both agents choose action $l_{13}$ at node $n_1$ simultaneously or one agent chooses action $l_{13}$ first because she arrives at $n_1$ first and the other agent chooses action $l_{13}$ later due to her later arrival at node $n_1$ (e.g., one agent chooses route $n_0 \rightarrow n_1$ and arrives at $n_1$ at $t = 45$, and the other agent chooses route $n_0 \rightarrow n_2 \rightarrow n_1$ and arrives at $n_1$ at $t = k_{02} + \alpha = 50$). In the former case, both agents observe $\bar{a} = 2$ and spend time $80$ on their way to $n_3$; in the latter case, the agent who arrives at $n_1$ first observes $\bar{a} = 1$ and spends time $40$ on her way to $n_3$ and the other agent observes $\bar{a} = 2$ and spends time $80$. With $Q(n_1,l_{13},1) = -40$ and $Q(n_1, l_{13}, 2) = -80$, optimal value at $n_1$, i.e., $V(n_1)$, is within the range of $[-40,-80]$ and depends on the distribution of $\bar{a}$. 
			\item At node $n_2$, there are two outbound links, namely $l_{21}$ and $l_{23}$. If one or two agents at $n_2$ choose action $l_{21}$, $Q(n_2,l_{21}, \bar{a}) = -\alpha + \gamma \times V(n_1) = -10 + V(n_1)$, which is lower than $-50$, regardless of $\bar{a}$. If one or two agents at $n_2$ choose action $l_{23}$, $Q(n_2, l_{23}, \bar{a}) = -\Delta t_{23} = -45$. Therefore, action $l_{21}$ is strictly dominated by action $l_{23}$, indicating that optimal policy at $n_2$ for both agents is to choose action $l_{23}$ and $V(n_2) = -45$. 
			\item Finally, at node $n_0$, there are two possible actions, namely $l_{01}$ and $l_{02}$, and both agents choose action simultaneously. There are 3 possible scenarios: 1) when both agents choose action $l_{01}$, $Q(n_0, l_{01}, \bar{a} = 2) = -\Delta t_{01} + \gamma V(n_1) = -45 + (-80) = -125$. Note that $V(n_1) = -80$ because both agents arrive at node $n_1$ and choose action $l_13$ at the same time. 2) when both agents choose action $l_{02}$, $Q(n_0, l_{02}, \bar{a} = 2) = -\Delta t_{02} + \gamma V(n_2) = -80 +(-45) = -125$. 3) when one agent chooses action $l_{01}$ and the other chooses action $l_{02}$, $Q(n_0,l_{01},\bar{a} = 1) = -\Delta t_{01} +\gamma V(n_1) = -45 + (-40) = -85$ and $Q(n_0, l_{02}, \bar{a} = 1) = -\Delta t_{02} + \gamma V(n_2) = -40 + (-45) = -85$. Note that $V(n_1) = -40$ because only one agent uses link $l_{13}$.  Therefore, the optimal policy of one agent is to choose action $l_{01}$ and that of the other agent is to choose action $l_{02}$ at $n_0$. %, and no agent can get better payoff by a unilateral deviation. 
		\end{enumerate}
		
		In summary, optimal values of interest are listed in Table~\ref{tab:vals}. Link $l_{21}$ is not used by agents because it is strictly dominated by $l_{23}$. From $n_0$, the payoff for agents are $(-85,-85)$ if they choose actions differently and $(-125,-125)$ if they choose the same action (i.e., either $l_{01}$ or $l_{02}$). Therefore, optimal policy of one agent is to choose action $l_{01}$ and that of the other is to choose $l_{02}$, and this is a Nash equilibrium because no agent can get better payoff by a unilateral deviation. 
		
		\begin{table}[H]
			\centering\caption{Optimal values of interest}
			\label{tab:vals}
			\begin{tabular}{ |p{60pt}| p{40pt}| p{60pt} | p{40pt}| p{180pt}|} 
				\hline
				observation $o$ & action $a$ &  mean action $\bar{a}$ & $Q(o,a,\bar{a})$ & $V(o)$\\ \hline
				$n_1$ & $ l_{13}$  & $1$ & $-40$ & \multirow{2}{180pt}{$V(n_1) = \mathbb{E}_{\bar{a}}[Q(n_1,l_{13}, \bar{a})]$ is within the range of $[-80,-40]$. 
					%Specifically, $V(n_1) = -40$ if $\bar{a} = 1$ and $V(n_1) = -80$ if $\bar{a} = 2$.
				}\\
				$n_1$ & $ l_{13}$  & $2$ & $-80$ & \\ \hline
				$n_2$ & $ l_{21}$  & $1$ or $2$ & $\leq -50$ & \multirow{2}{180pt}{Action $l_{23}$ strictly dominates $l_{21}$. $V(n_2) = Q(n_2,l_{23},\bar{a}) = -45$.}\\ 		
				$n_2$ & $ l_{23}$  & $1$ or $2$ & $-45$ &\\ \hline			
				$n_0$ & $ l_{01}$  & $1$ & $-85$ & \multirow{4}{180pt}{$V(n_0) = -85$ if agents choose actions differently at $n_0$ and $V(n_0) = -125$ if agents choose the same action. }\\ 		
				$n_0$ & $ l_{01}$  & $2$ & $-125$ &\\
				$n_0$ & $ l_{02}$  & $1$ & $-85$ &\\
				$n_0$ & $ l_{02}$  & $2$ & $-125$ &\\ \hline								
			\end{tabular}
		\end{table}
	\end{exmp}

Markov games to model en-route choice are independent of data-generating mechanism. 
Here data refers to background traffic in which vehicles move and by which travel choices are influenced, including but not limited to travel time or delay and/or travel speed on a link, queuing delay at a node, as well as transition flows between links.
%referred to as in- and out-flow, travel time or delay and/or travel speed on a link, as well as individuals' movement trajectories.} 
Routing flows can arise from models like DNL, or model-free simulation. 
In the next two sections, we will demonstrate the proposed routing game using data first generated from existing DNL models %(which is model-based) 
and then from a traffic simulator where its data generating mechanism remains unknown. %(which is model-free).

\section{Linkage between Dynamic traffic assignment (DTA) and Markov routing games (MRG)}\label{sec:DUE}

%\tcr{Xu, can you enrich lit. on DTA below, including those from Kuang's and others you recently reviewed?}

% \tcr{Xu/Zhenyu, maybe we should justify that MARL can solve MFG on networks more efficiently, such as Sioux Falls. Because so far it is super challenging for us to solve MFE on mid-size or large-size networks, with convergence issues. 
% If possible, Xu, can you try to generate an environment using MFG on each link (instead of CTM, including speed choice) for Zhenyu to solve routing.}

DTA is the descriptive modeling of dynamic traffic flows on networks that is consistent with traffic flow theory, travel behaviors, and established travel demands. 
It has been widely used to model route choice behavior of travelers. A general DTA problem typically consists of two parts, namely a dynamic network loading (DNL) module and a route choice model. 

\textbf{DNL}. Given traffic demand and route choices, the DNL module propagates traffic flow dynamics and/or traffic congestion through the network. There are two components in the DNL module, namely a link model and a junction model. A link model is used to propagate inflow traffic demand to exit and is usually described by either a delay/latency function or an exit-flow function \citep{nie2005comparative}. As for the former, both linear delay functions and nonlinear functions (e.g., the widely used BPR function) are used in the literature. As for the latter, there are a large body of literature proposing or using various exit-flow functions such as the M-N model \citep{merchant_model_1978,merchant_optimality_1978}, the point-queue model \citep{kuwahara_decomposition_1997,ban_continuous-time_2012}, and the cell transmission model \citep{daganzo_cell_1994, daganzo_cell_1995}. 
%Interested readers can refer to \cite{kuang} for more exit-flow models. 
A junction model is used to determine the flow distribution at an intersection. 

\textbf{Route choice}. Route choice models guide travelers to properly choose next go-to links right before they arrive at a node/intersection. According to the objective of travelers, there are typically two types of research problems, namely dynamic system optimum (DSO) and dynamic user equilibrium (DUE). While DSO aims to minimize total or average travel cost of all travelers \citep{ziliaskopoulos_linear_2000}, DUE describes the equilibrium state in which no individual traveler  could decrease her travel cost by unilaterally deviating from her current route choices \citep{friesz2019mathematical,friesz2013dynamic}. 
\cite{huang2021driving} applies mean-field game to model both velocity and route choices of a large number of intelligent agents on a road network.

\begin{figure}[H]
	\centering
	\includegraphics[width=0.99\linewidth,height=0.3\textheight,keepaspectratio]{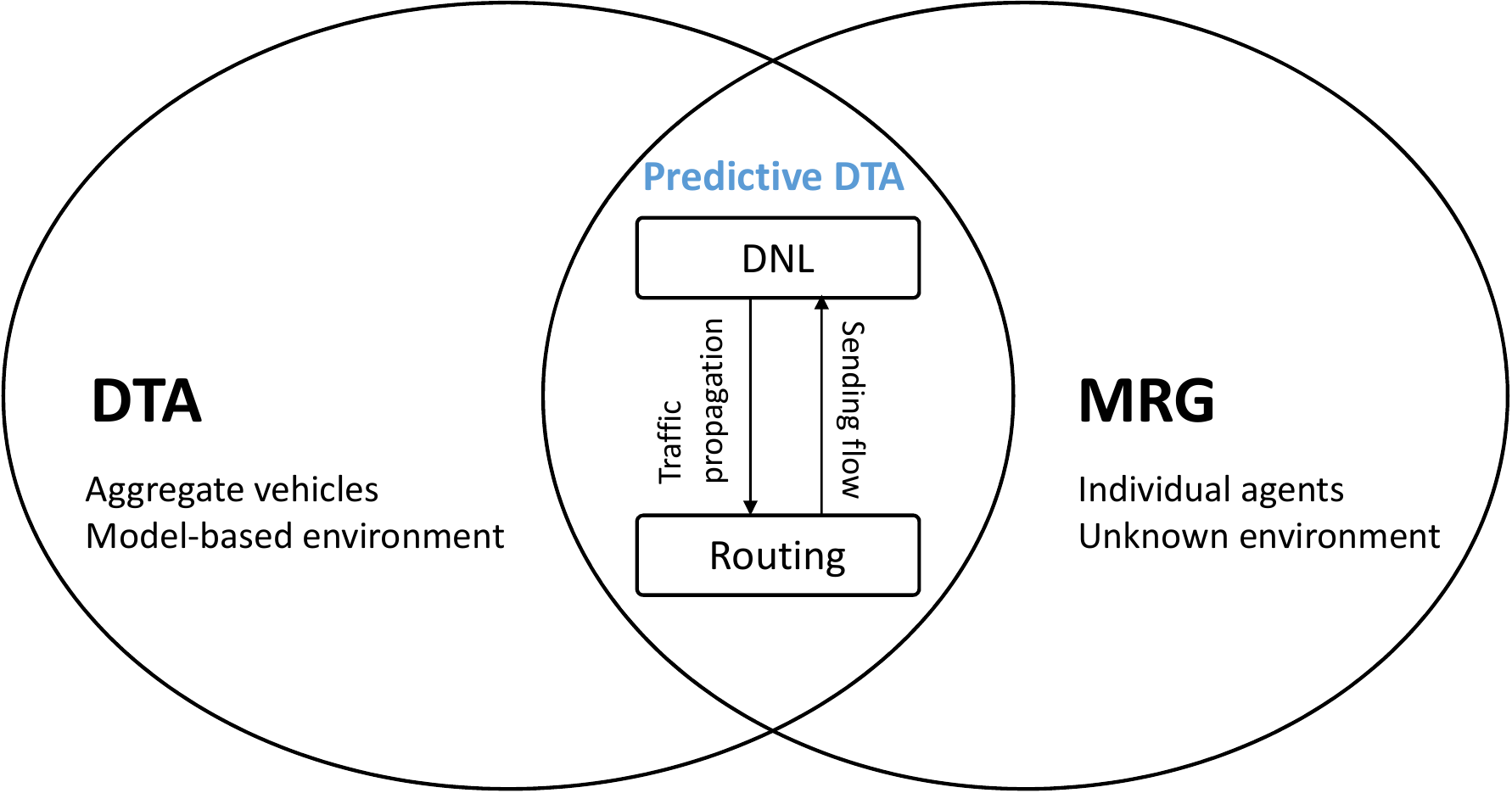}
	\centering 
	\caption{Connection between DTA and Markov routing game (MRG)}
	\label{fig:dta_mg}
\end{figure}

We now discuss the difference and connection between DTA and the MRG developed in this paper. 
Deriving a Nash equilibrium in general requires a perfect knowledge of the system evolution dynamics and the rewards of the game \citep{perolat2017learning}. 
However, the proposed MRG models individual agents' routing behavior with a partial or unknown information structure and a stochastic model-free environment. 
In other words, in MRG, agents lack the knowledge of the transition probability of system dynamics or travel time functions of road segments. 
Thus, agents need to learn through repeated interactions with other agents and the traffic environment. 
This setting mimics the reality that we usually have no knowledge of how traffic evolves nor a mathematical relation between travel time and congestion. 
This is the key difference between DTA models and MRG presented in this paper. 
DTA models stipulate the system evolution dynamics and the travel cost of using each link or route, while MRG does not contain such information and thus requires agents to learn and adapt their strategy according to historical episodes generated from interactions.

Other than the difference between DTA and MRG, we would like to draw their linkage. 
Markov game is the generalization of MDP-like environment to a multi-agent system. 
The environment of an agent is comprised of other agents and unknown objects. In other words, stochasticity of the environment arises from the actions of other agents and unknown sources. 
To generate such an MDP-like environment, we not only need to model all the agents' Markov strategies, but also their interactions with one another.
Various DNL models of DTA can provide such an environment. Specifically, there are two components in a DTA, namely a DNL module and a route choice model, and two components in a Markov game, namely an environment and agents. 
The connection between DTA and Markov game is componentwise, as schematically presented in Figure~(\ref{fig:dta_mg}). 
The DNL in DTA can be used as the environment engine in Markov game, 
while the route choices in DTA are made by agents in Markov game. 
Further, the interaction between the DNL and route choice in DTA is similar to that between the environment and agents in Markov game. In DTA, a route choice model outputs route choices of travelers whenever they arrive at a node or an intersection, and the DNL module propagates traffic flow dynamics. 
In MRG, agents (i.e., travelers) take actions (i.e., route choices) whenever they arrive at a node, 
and the environment experiences a state transition determined by the DNL module in DTA. To be precise, the actions taken by agents in Markov game are route choices in DTA, and the state transition and rewards in Markov game are determined by the DNL module in DTA. 
Moreover, DUE and MRG have non-overlapped regimes that delineate the differences of these two model types: 
DTA models encompass those with reactive DUE \citep{li2000rdue,lam1995rdue} and are primarily solved using aggregate traffic flow variables; 
while MRG is focused on en-route dynamic choice of individuals within traffic environments of which transition dynamics and travel time functions remain unknown to agents.

%\textcolor{blue}{(Markov game should enclose DTA to echo my statement below.) As pointed out in the figure, DUE is essentially a special case of multi-agent Markov games with a perfect information structure and a deterministic environment described by mathematical models.}

In summary, 
the MRG coupled with DNL models depicts one type of DTA models, of which the decision-making processes of discrete vehicles (in other words, atomic agents \citep{roughgarden2007routing}) instead of aggregate traffic flows are modeled, 
traffic dynamic evolution is modeled as a stochastic transition probability without any need of model specification and in discrete time steps, 
and the equilibrium outcome is predictive DUE.
The detailed comparisons of DTA and MRG are summarized in Table~\ref{tab:dta_mg}.
\begin{table}[H]\centering
	\caption{DTA versus MRG}\label{tab:dta_mg}
\begin{tabular}{p{.8 cm}||p{1.1 cm}|p{1.5 cm}|p{1.4 cm}|p{1.5 cm}|p{1.3 cm}|p{1.5 cm}|p{1.3 cm}|p{1.3 cm}}
	\hline
	\multirow{2}{*}{Model} & Game & \multicolumn{4}{|c|}{Traffic dynamics} & \multicolumn{2}{|c|}{Travel Cost} & \multirow{2}{*}{Equilibrium} \\\cline{2-8}
	& Atomic or non-atomic & Knowledge of dynamics & State transition & Functional form & Temporal resolution & Knowledge of reward & Form &  \\ \hline\hline
	DTA & Non-atomic & Perfect & Stochastic or deterministic & Model-based & Discrete or continuous & Known & Actual or instantaneous & Predictive or reactive \\ \hline
	MRG & Atomic & Imperfect & Stochastic & Model-free & Discrete & Unknown & Actual & Predictive \\ \hline
\end{tabular}
\noindent Atomic -- Discrete vehicles; Non-atomic -- Aggregate traffic flow.
\end{table}

We further specify two components, namely, dynamic network loading and route choice, in DTA and MRG, respectively, using mathematical representations.  
%Note that the linking variables from individual agents to aggregate traffic flows are explained.   

\noindent\fbox{
	\parbox{0.97\textwidth}{
DTA: find optimal path flow $h^{t*}_k,k \in \mathcal{K}$, $\forall t\in \{0, T\}$.
\begin{itemize}
	\item Dynamic network Loading: 
	$\begin{cases}
	    N_{l}^{\uparrow}(t+1)= N_{l}^{\uparrow}(t)+\sum_{l'\in \Gamma^{-1}(l)} y_{l'l}(t), \\
	    N_{l}^{\downarrow}(t+1)= N_{l}^{\downarrow}(t)+\sum_{l^{''}\in \Gamma(l)} y_{ll''}(t),
	\end{cases}$
	$l\in\mathcal{L}^D,\, 0\leq t\leq N_t-1,$
	where $N^{\uparrow}_{l}$ and $N^{\downarrow}_{l}$ are the cumulative counts of vehicles at the upstream and downstream of link $l$, respectively. $\sum_{l'\in \Gamma^{-1}(l)} y_{l'l}$  and $\sum_{l^{''}\in \Gamma(l)} y_{ll''}$ are the inflow and outflow of link $l$, respectively. 
	
	%$x_{l}^{t+1}=x_{l}^t+\sum_{l'\in \Gamma^{-1}(l)} y_{l'l}^t - \sum_{l^{''}\in \Gamma(l)} y_{ll''}^t,\quad l\in\mathcal{L}^D,\, 0\leq t\leq N_t-1$; $x_{l}^t$ is the number of vehicles on link $l$. 
	
  \item Route choice for aggregate vehicle flows:
	
$\forall k \in \mathcal{K}$,
$\begin{cases} 
    h^t_{k}>0 \rightarrow 
    \pi^t_{k} = min_{k} \mathcal{T}(H),
    \nonumber\\
    h^t_{k}=0 \rightarrow
    \pi^t_{k} > min_{k} \mathcal{T}(H), \label{eq:dta} \nonumber
\end{cases}$

where, $h^t_{k}$ is the traffic flow on path $k \in \mathcal{K}$, $H$ is the path flow matrix, $\mathcal{T}(H)$ is the travel time matrix regarding each path and $\pi^t_{k}$ is the minimum travel time.
\begin{rem}
 \begin{enumerate}
     \item $y_{l'l}$ is the transit flow from link $l'$ to link $l$, depending on sending and receiving flow in DNL models, which will be introduced in Section 4.
     \item Each element $\tau^t_{k}$ in $\mathcal{T}(H)$ is the travel time on path $k$ and $\tau^t_{k} =T^{\downarrow}_{k}(N^{\uparrow}_{k})-T^{\uparrow}_{k}(N^{\uparrow}_{k})$. $T^{\downarrow}_{k}$ and $T^{\uparrow}_{k}$ represent the time when vehicles leave and enter path $k$, respectively. 
 \end{enumerate}
\end{rem}
% \item Flow conservation:
% 	$p_{ij}^t=\beta_{ij}^t\left(\sum_{(k,i)\in\mathcal{L}^D}q_{ki}^t + d_i^t - \frac{Q_i^{t+1}-Q_i^t}{\Delta t}\right)$,\quad $(i,j)\in\mathcal{L}^D$,\, $0\leq t\leq N_t-1$;\label{eq:micp3}
% \item Link travel time function: $\forall (i,j)\in L$,
%     $\tau_{ij}(t) = \tau_{ij}(\rho_{ij}(t))$
\end{itemize}
	}
}

\noindent\fbox{
\parbox{0.97\textwidth}{
MRG: find optimal policies $\mu^{*}_i, i=1, \cdots, N$, $\forall t\in \{0, T\}$.
\begin{itemize}
	\item Environment transition: 
 $P(\mathbf{s'}|\mathbf{s}, \mathbf{a})$, where $\mathbf{s}$ represents the environment state, i.e., the agent distribution on the network, which determines the cumulative counts of vehicles $N_l^{\uparrow}, N_l^{\downarrow},l \in \mathcal{L}^D$. 
	\item Route choice for each agent: 
	
	$\forall i\in N$,
$\begin{cases} 
    a^{t}_{i,l}>0 \rightarrow 
    V_i(s) = max_{a_i} 
	\mathbb{E}_{\mathbf{a}^{-i}\sim A}
	\mathbb{E}_{s'\sim P(\cdot|s,\mathbf{a})}[r_i(s,\mathbf{a},s')+\gamma V_i(s')], \nonumber\\
    a^t_{i,l}=0 \rightarrow 
    V_i(s) < max_{a_i} 
	\mathbb{E}_{\mathbf{a}^{-i}\sim A}
	\mathbb{E}_{s'\sim P(\cdot|s,\mathbf{a})}[r_i(s,\mathbf{a},s')+\gamma V_i(s')]. \label{eq:mrg} \nonumber
\end{cases}$

where, for agent $i$, $a^t_{i,l}$ is her route choice, indicating whether agent $i$ chooses link $l$ or not. The reward $r_i(s,\mathbf{a},s')$ is the negative realized travel time after agent $i$ choosing action $a_i$.
		
% \item Flow conservation:
% $\beta_{ij}^t = \sum_{i} 1\{a^{t(i)}_{ij}>0\}$

% 	$p_{ij}^t=\beta_{ij}^t\left(\sum_{(k,i)\in\mathcal{L}^D}q_{ki}^k+d_i^t-\frac{Q_i^{t+1}-Q_i^t}{\Delta t}\right)$,\quad $(i,j)\in\mathcal{L}^D$,\, $0\leq t\leq N_t-1$;\label{eq:micp3}

% \item Link propagation dynamics: 
%      $x_{ij}^t=\rho_{ij}^t u_{ij}^t$,\quad $(i,j)\in\mathcal{L}^D$,\, $0\leq t\leq N_t-1$	
\end{itemize}
}
}

\section{Numerical examples}\label{sec:case}

In this section, we demonstrate the effectiveness of our MF-MA-DQL algorithm to solve optimal routing policies. In Example 4.1, we apply the developed algorithm to a simple network based on a variety of DNL models and compare it with a traditional iterative method. In Example 4.2, we specify one of the DNL models - LTM to explore the spillback phenomenon. In Example 4.3, we apply the developed algorithm to the OW Network \citep{ow2011network} with LTM as its data-generation environment. In Example 4.4, our algorithm is applied to a real-world network with traffic simulated in SUMO (i.e., model-free environment). 
To reiterate, the first three examples contain model-based traffic environments using DNL models, while the last one has a model-free environment in which travel time function and traffic state update are unknown and thus requires agents to learn from their experiences.

\begin{table}[H]\centering
	\caption{Numerical examples}\label{tab:numerical_example}
	\begin{tabular}{c|c|c|c}
		\hline
		& Example &Network & Dynamic Loading Environment \\ %[5pt]
		\hline
		\multirow{3}{*}{Model-based} & 4.1  & Simple & \makecell{PQ,\
		 SQ,\ CTM,\ LTM}\\[5pt] \cline{2-4}
		&4.2 & Simple (spillback) & LTM \\[5pt]  \cline{2-4}
		&4.3 & OW & LTM \\[5pt]  \hline
		Model-free &4.4  & Real-world network &  SUMO \\[5pt] \hline
	\end{tabular}
\end{table}

\begin{exmp}\label{exm:DSO_simple}
In this example, we apply the developed MF-MA-DQL algorithm to a simple network. We first briefly introduce several DNL models used as the environment of MARL.
\begin{enumerate}
    \item \textbf{PQ:} In the point queue (PQ) model, vehicles have no physical length (no spillback) and move at free flow speed. The link traversal time consists of free flow travel time and queuing delay \citep{zhang2013pq}.
    \item \textbf{M-N:} In the M-N model, exit flow is a generalized function of road density \citep{merchant_model_1978}. When the function satisfies flow capacity constraints, the M-N model becomes a special case of PQ \citep{nie2011m_n}. For simplicity, we use the PQ model to demonstrate our algorithm.
    
    \item \textbf{SQ:} The different between spatial queue (SQ) and PQ models is that the exit flow in the SQ model depends on the flow capacity and the remaining capacity of downstream links \citep{zhang2013pq}.
    
    \item \textbf{CTM:} The cell transmission model (CTM) approximately solves the LWR system by discretizing space and time \citep{daganzo_cell_1994, daganzo_cell_1995}. CTM considers not only flow capacity and jam density, but also backward speed into the fundamental diagram.
    
    \item \textbf{LTM:} The link transmission (LTM) model exactly solves the LWR system using the Newell-Daganzo method \citep{yperman2007LTM}. Compared with CTM, LTM reduces computational complexity.
    
    \item \textbf{DQ:} The deterministic double queue (DQ) model coincides with LTM \citep{caro2011dq}. For simplicity, we use LTM in this paper.
\end{enumerate}

We then show how to construct the environment of MARL based on DNL models. Note that a link can be discretized into cells such that the length of each cell is exactly the distance traveled by a vehicle within a time interval $\Delta t$ at free flow speed. We assume that the time interval $\Delta t$ is also the time step in the environment of our MARL. To better understand this, we use LTM to illustrate the environment. Environments based on other DNL models are in \ref{append:DNL_environment}.

LTM Environment: LTM is applied to only to the upstream and downstream ends of a link \citep{yperman2007LTM,caro2011dq}. It is a discrete model, calculating the upstream and downstream flow at each time step. At time $t$, the cumulative counts of vehicles at the upstream and downstream ends of the link are denoted by $N^{\uparrow}(t)$ and $N^{\downarrow}(t)$, respectively.  The sending flow $S(t)$ at the downstream end of the link is calculated based on the free flow speed $v$, representing how many vehicles can leave during the time interval $(t,\ t+\Delta t)$, which is calculated as: 
\begin{align}
    S(t)=min\{N^{\uparrow}(t-\frac{L}{v}+\Delta t)-N^{\downarrow}(t),\ q_{max} \Delta t \}, \nonumber 
\end{align}
where $L$ represents the link length and  $N^{\uparrow}(t-\frac{L}{v}+\Delta t)$ represents the upstream cumulative count of vehicles at time $t-\frac{L}{v}+\Delta t$. $q_{max}$ represents the flow rate capacity and $q_{max} \Delta t$ is the maximum number of vehicles that can leave at each time step. The receiving flow $R(t)$ at the upstream end of the link is calculated based on the congested wave speed (backward speed) $w$, representing how many vehicles enter the link during the time interval $(t,\ t+\Delta t)$, which is calculated as: 

\begin{align}
    R(t)=min\{N^{\downarrow}(t-\frac{L}{w}+\Delta t)+k_j L-N^{\uparrow}(t),\ q_{max} \Delta t \}, \nonumber 
\end{align}
where $N^{\downarrow}(t-\frac{L}{w}+\Delta t)$ is the downstream cumulative count of vehicles at time $t-\frac{L}{w}+\Delta t$, $k_j$ is the jam density and $k_j L$ is the link capacity, representing the maximum number of vehicles that can be accommodated into the link.

Note that the receiving and sending flows only capture the upstream and downstream ends of a link. To determine the flow propagation in the environment, we also need node models to calculate how many vehicles can move from an upstream link to a downstream link. There are some simple node models capturing merging and diverging cases \citep{daganzo_cell_1994}. However, general intersection with more than one inflow link and one outflow link is not standardized as merge and diverge models. It can be signal-controlled or priority-controlled. In this paper, we use discharging priorities \citep{ma2014dq} based on agents' route choice at a general intersection node where the inflow links of the node are ranked in some order, representing different priorities. Vehicles on each inflow link may leave based on the link priority and enter outflow links based on their route choice (actions).

% \textcolor{blue}{Take the PQ model with flow capacity $\bar{q}=2/h$ as an example.  There is a link $a\rightarrow b$ ($a$ is the entry node and $b$ is the exit node of the link) where link length is 10 mile and free flow speed is 5 mile/h. Accordingly, the time step in the environment is $1 h$ and the link is divided into 2 cells: $c_1,c_2$. At the time step $t=0$, four vehicles with an order $v_1,v_2,v_3,v_4$ start entering cell $c_1$ from node $a$. The flow capacity of PQ is $2/h$,  which means in a time step, the first two vehicles $v_1,v_2$ can enter cell $c_1$. At the next time step $t=1$, vehicles $v_1,v_2$ in cell $c_1$ enter cell $c_2$ and $v_3,v_4$ enter cell $c_1$. When $t=2$, vehicles $v_1,v_2$ reach node $b$ and $v_3,v_4$ enter cell $c_2$. When $t=3$, all vehicles reach node $b$.}

With the aforementioned dynamic loading environment, now we apply the developed MF-MA-DQL algorithm to solve DUE where travel cost on all used routes for any given origin-destination pair is supposed to be the same according to Wardrop's first principle. At equilibrium, no individual selfish vehicle could decrease her travel cost by unilaterally deviating from her current route choice strategy. Therefore, the reward for each is simply the negative of her own travel time. In other words, each selfish agent aims to minimize her own travel time.

\begin{figure}[H]
	\centering 
	\includegraphics[scale=.4]{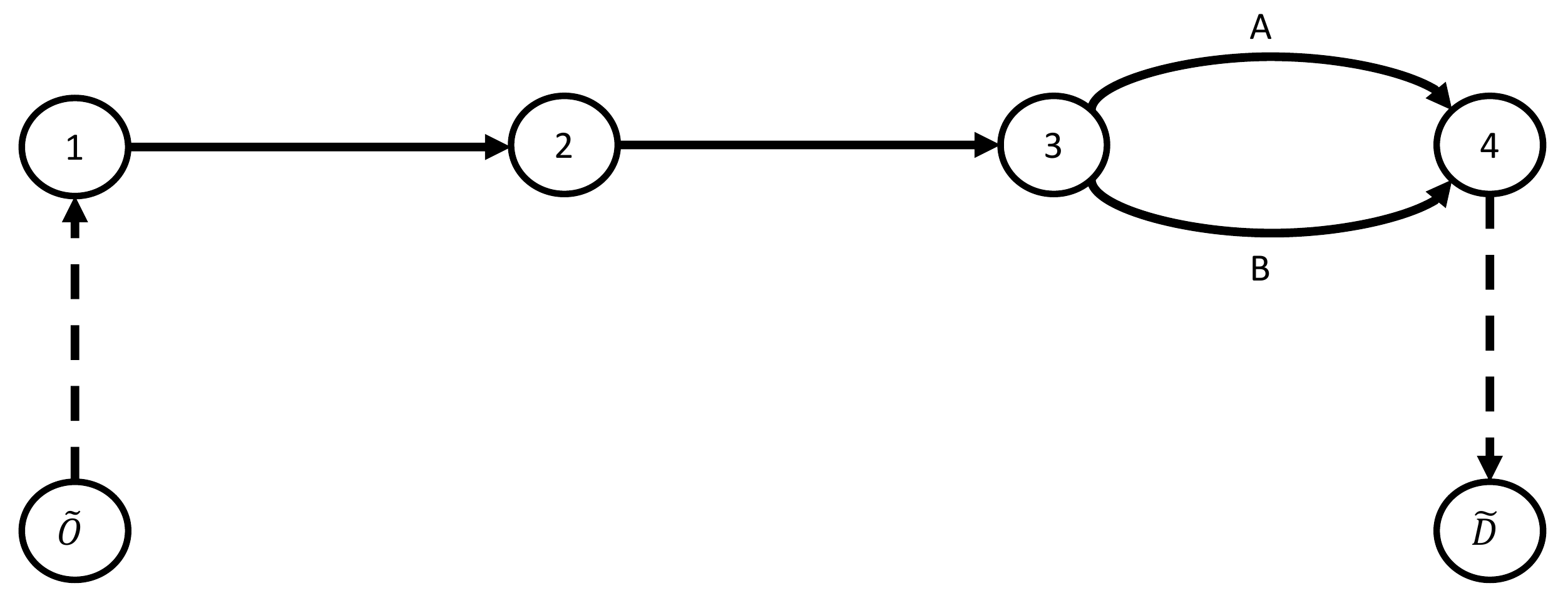}
	\caption{Simple Network}
	\label{fig:simple_net} 
\end{figure}

The simple network used to test the proposed algorithm is shown in Figure \ref{fig:simple_net}. The origin and destination are node 1 and 4, respectively. When vehicles reach node 3, they have to make a route choice between link $3\rightarrow A \rightarrow 4$ and  $3\rightarrow B \rightarrow 4$. $\Tilde{O}$ is the dummy node of origin. We assume that at the beginning, vehicles are on the dummy link $\Tilde{O}\rightarrow 1$, representing actual travel demand at the origin node \citep{ma2014dso}.  $\Tilde{D}$ is the dummy node of destination. All vehicles are on the dummy link $ 4\rightarrow \Tilde{D}$ after reaching destination. The travel demand at the origin node $d(t)$ is $d(0)=50$, which means there are 50 vehicles on the dummy link $\Tilde{O}\rightarrow 1$ at time 0. The free flow speed is $v=0.2$. Other parameters of this network is shown in Table \ref{tab:para_simple_4.1} where link length, flow rate capacity, jam density and backward speed are denoted as $L$, $q_{max}$, $k_j$, and $w$, respectively. Note that SQ, LTM and CTM share the same link length and flow capacity with PQ. LTM and CTM share the same link capacity with SQ.

\begin{table}[H]
  \centering
  \begin{tabular}{c|c|c|c|c}
    \hline
    \multirow{2}{*}{Link} &
      \multicolumn{2}{c|}{PQ} &
      \multicolumn{1}{c|}{SQ} &
      \multicolumn{1}{c}{LTM/CTM} 
      \\ \cline{2-5} 
    & $L$ & $q_{max}$ & $k_j$  & $w$\\
    \hline
    $\Tilde{O} \rightarrow 1$ & $\infty$ & $\infty$ & - & -   \\
    \hline
    $1 \rightarrow 2$ & 0.4   & 20 & 300 & 0.1 \\
    \hline
    $2 \rightarrow 3$ & 0.4   & 20 & 300 & 0.1 \\
    \hline
    $3 \rightarrow A  \rightarrow4$   & 0.8 & 2 & 60 & 0.1 \\
    \hline
    $3 \rightarrow B  \rightarrow 4$   & 0.4 & 2 & 30 & 0.1 \\
    \hline
    $4 \rightarrow \Tilde{D}$   & $\infty$ & $\infty$ & - & - \\
    \hline
  \end{tabular}
  \caption{Parameter in Example 4.1}
  \label{tab:para_simple_4.1}
\end{table}

We make a comparison of the developed algorithm and a traditional iterative method (See \ref{append:iteration_method}). We plot the convergent performance based on environments of PQ, SQ, LTM and CTM. Convergence plots of both methods for the DUE scenario are presented in Figure~\ref{fig:conv_due}. The x-axis represents the number of iterations/episodes and the y-axis is the average travel time. In Figure \ref{subfig:ite}, the iterative method reaches convergence when the average travel times on both links become the same. After $40$ iterations, the average travel times on both links $3-A-4$ and $3-B-4$ converge. The converged average travel time for $d(0)=50$ is $13.7$. At the equilibrium, there are $20$ and $30$ vehicles choosing link $3-A-4$ and $3-B-4$, respectively.

For the MF-MA-DQL algorithm, after bouncing back and forth in Figure~(\ref{subfig:due}), average travel travel time gradually reaches its converged value $13.7$, which is consistent with the iterative method. In this case, no spillback happens and travel times in all DNL environments converge to the same value.

\begin{figure}[H]
	\centering 
	\subfloat[The iterative method]{\includegraphics[scale=.4]{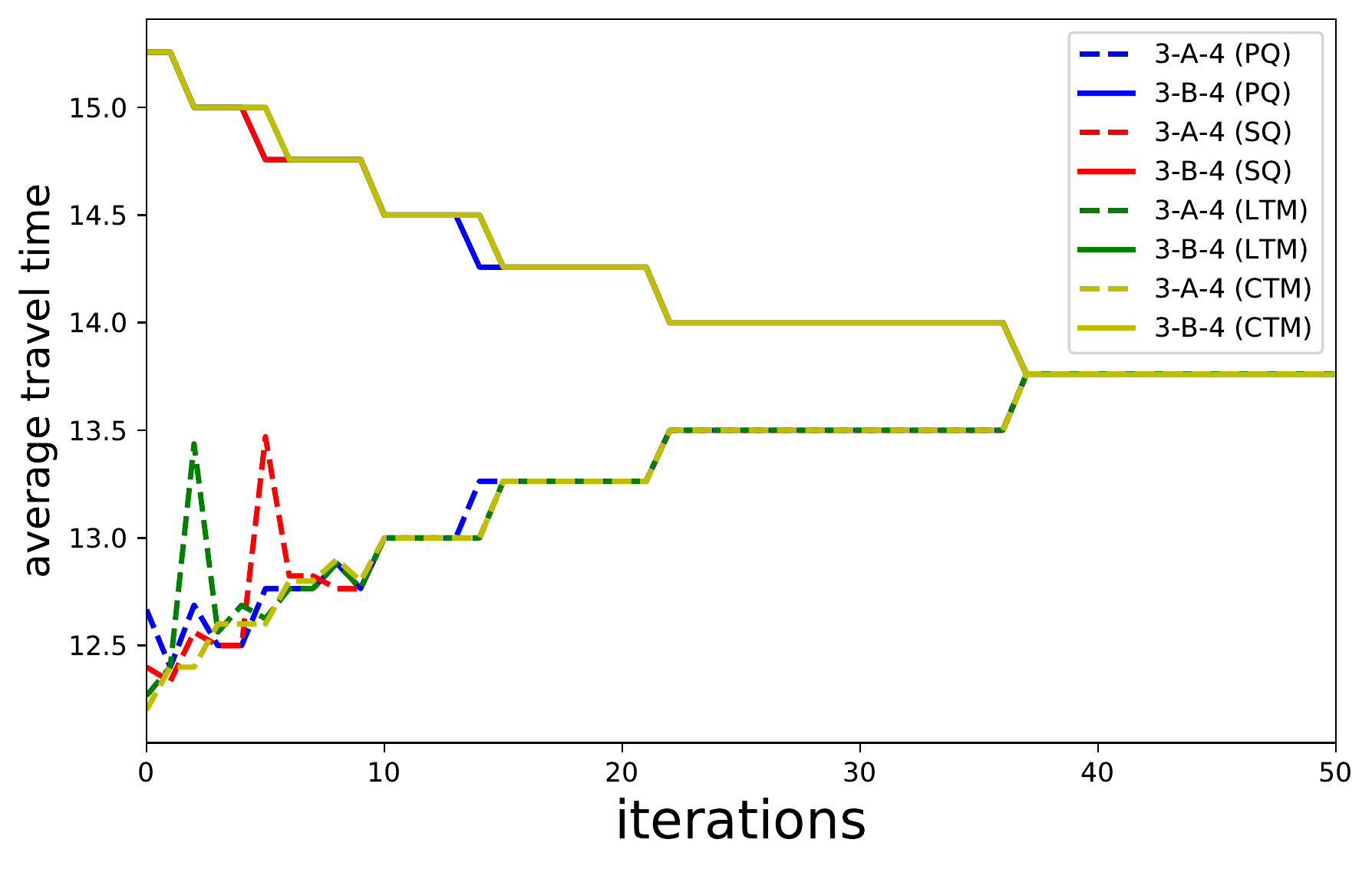}\label{subfig:ite}}
	\subfloat[MF-MA-DQL algorithm for DUE]{\includegraphics[scale=.4]{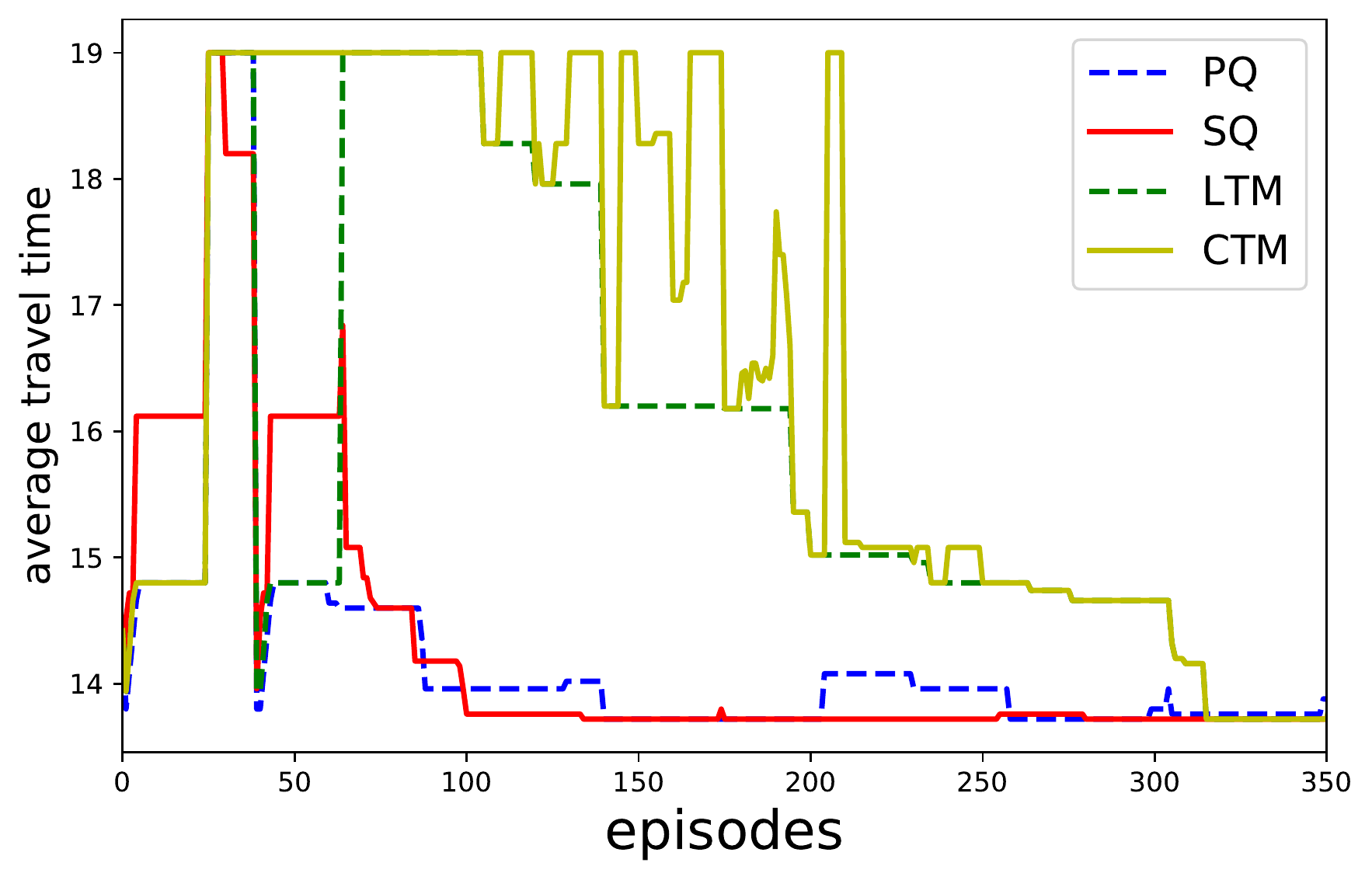}\label{subfig:due}}
	\caption{Convergence plots for DUE}
	\label{fig:conv_due} 
\end{figure}

\end{exmp}

\begin{exmp}\label{due_simple_spill}
In this example, we investigate a special case on the simple network where queue spillback may happen. The dynamic loading environment we use is LTM. The travel demand at the origin node $d(t)$ is $d(0)=90$, which means there are 90 vehicles on the dummy link $\Tilde{O}\rightarrow 1$ at time 0. The free flow speed is $v=0.2$ and the backward speed is $w=0.1$. The length of link $1 \rightarrow 2$, $2 \rightarrow 3$, $3 \rightarrow A \rightarrow 4$, $3 \rightarrow B \rightarrow 4$ are 0.2, 0.2, 0.4 and 0.4, respectively. The flow rate capacity of link $1 \rightarrow 2$ and $2 \rightarrow 3$ is 8. The flow rate capacity of dummy link $\Tilde{O}\rightarrow 1$ is 4, which means at each time step, four vehicles can leave the origin node. The flow rate capacity of  link $3 \rightarrow A \rightarrow 4$ and $3 \rightarrow B \rightarrow 4$ is:
\begin{equation}
   q_{max}=\left\{
\begin{aligned}
2, \ & t \leqslant 5 \\
1, \ & t>5
\end{aligned}
\right. \nonumber
\end{equation}

Convergence plots of both methods for the DUE scenario are presented in Figure~\ref{fig:spill_due}. In Figure \ref{subfig:ite_spill}, the iterative method reaches convergence when the average travel times on both links become the same. After $80$ iterations, the average travel times on both links $3-A-4$ and $3-B-4$ converge. The converged average travel time for $d(0)=90$ is $26$.

For the MF-MA-DQL algorithm, after bouncing back and forth during the first $200$ episodes in Figure~(\ref{subfig:due}), average travel travel time gradually reaches its converged value $26$, which is consistent with the iterative method. At the equilibrium, there are $45$ and $45$ vehicles choosing link $3-A-4$ and $3-B-4$, respectively.

Figure \ref{fig:queue} plots the queue spillback on the network for DUE. The x-axis represents the time and the y-axis represents the queue length. The queue length $q^{\downarrow}(t)$ at each time step is the number of vehicles in the downstream queue of each link \citep{ma2014dso,caro2011dq,yperman2007LTM} in the LTM environment, which is calculated as:

\begin{align}
    q^{\downarrow}(t)=N^{\uparrow}(t-\frac{L}{v}+\Delta t)-N^{\downarrow}(t). \nonumber 
\end{align}

It is shown that the queue forms on the link $2-3$ when $t=12$ due to the limited link capacity of its downstream links $3-A-4$ and $3-B-4$. After that, the queue moves backward and forms on the link $1-2$ when $t=20$ due to the limited link capacity of its downstream link $2-3$. 

\begin{figure}[H]
	\centering 
	\subfloat[The iterative method]{\includegraphics[scale=.43]{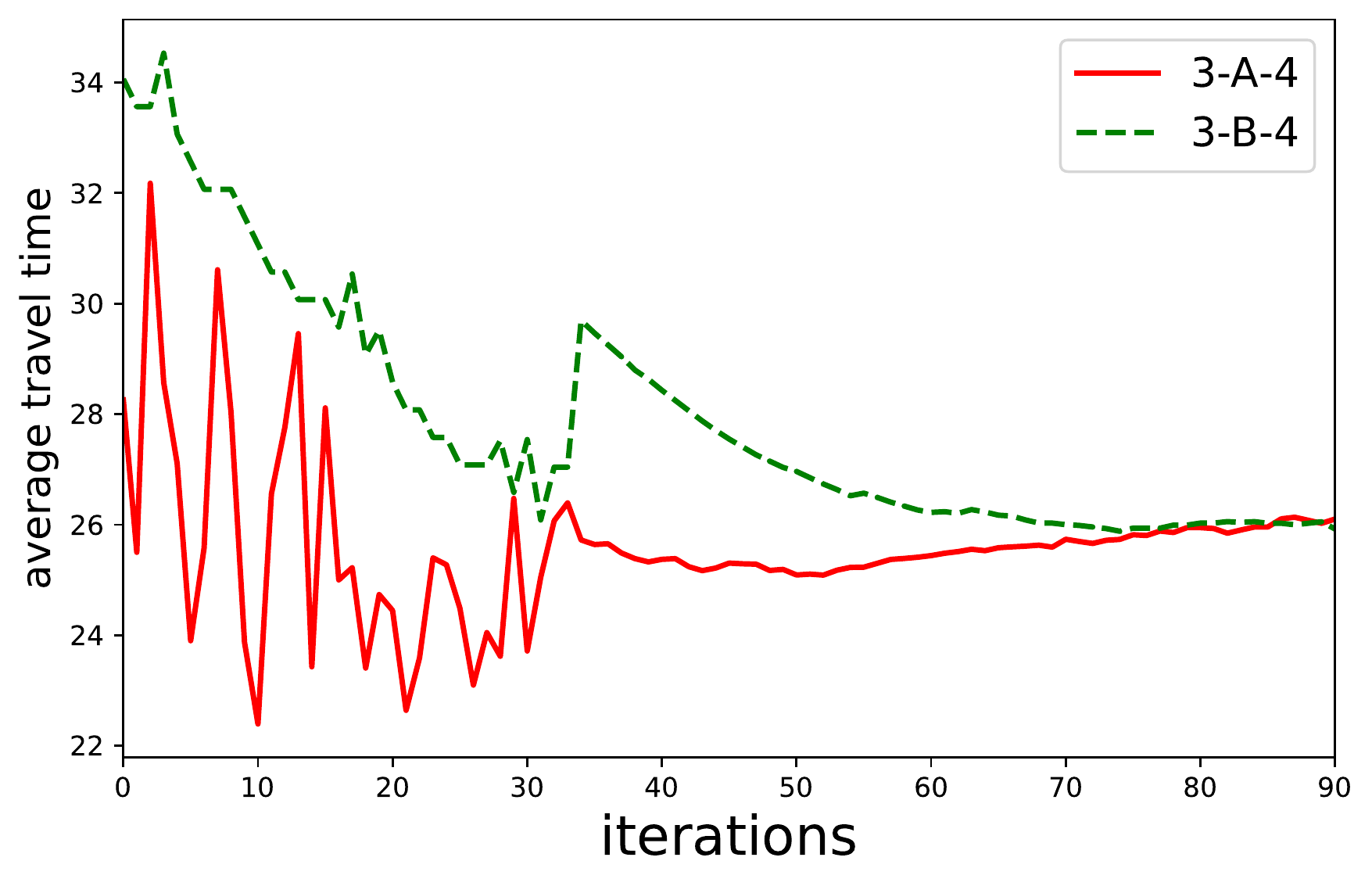}\label{subfig:ite_spill}}~~~
	\subfloat[MF-MA-DQL algorithm for DUE]{\includegraphics[scale=.43]{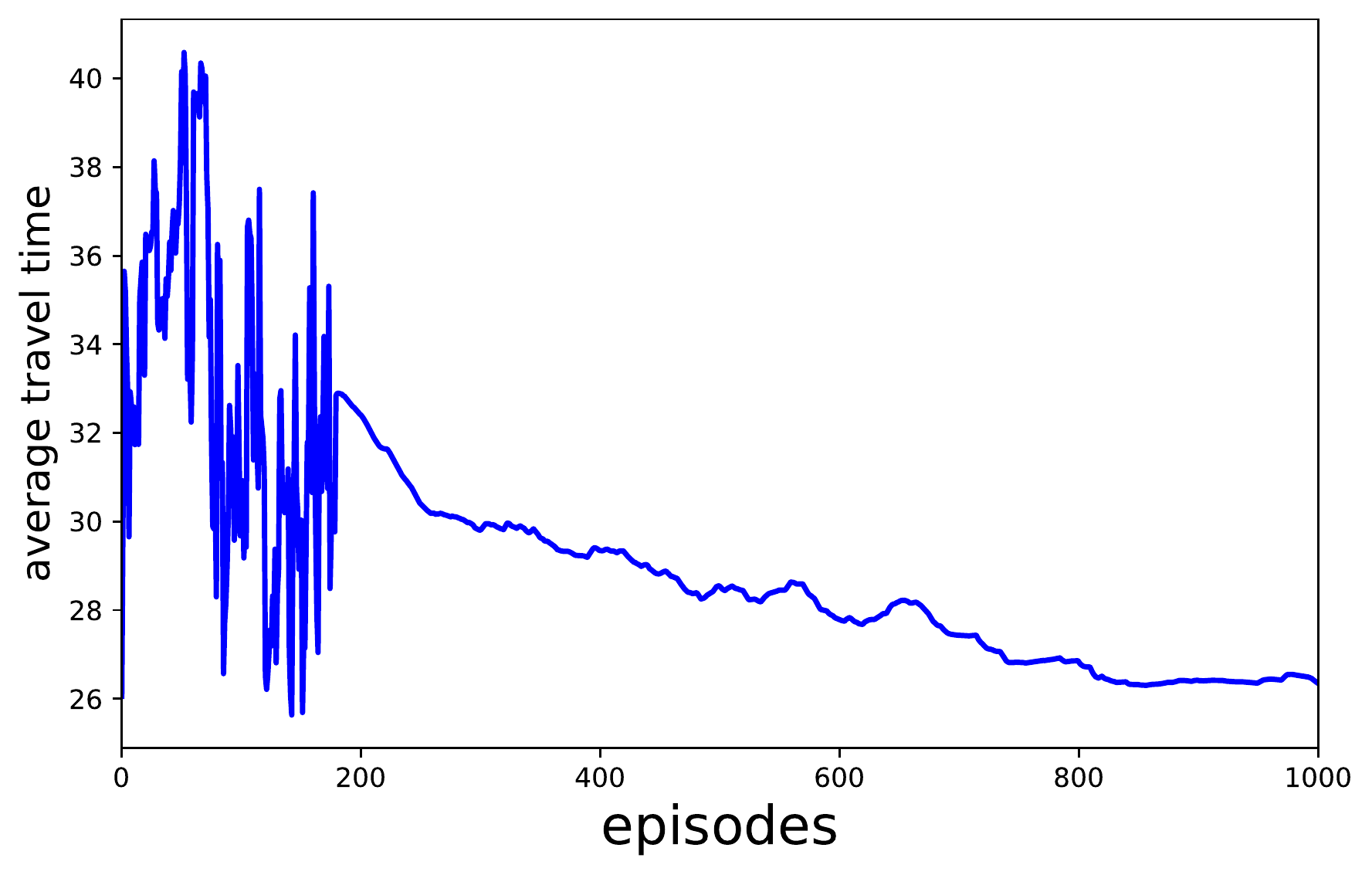}\label{subfig:due_spill}}
	\caption{Convergence plots for DUE}
	\label{fig:spill_due} 
\end{figure}

\begin{figure}[H]
	\centering 
	\includegraphics[scale=.43]{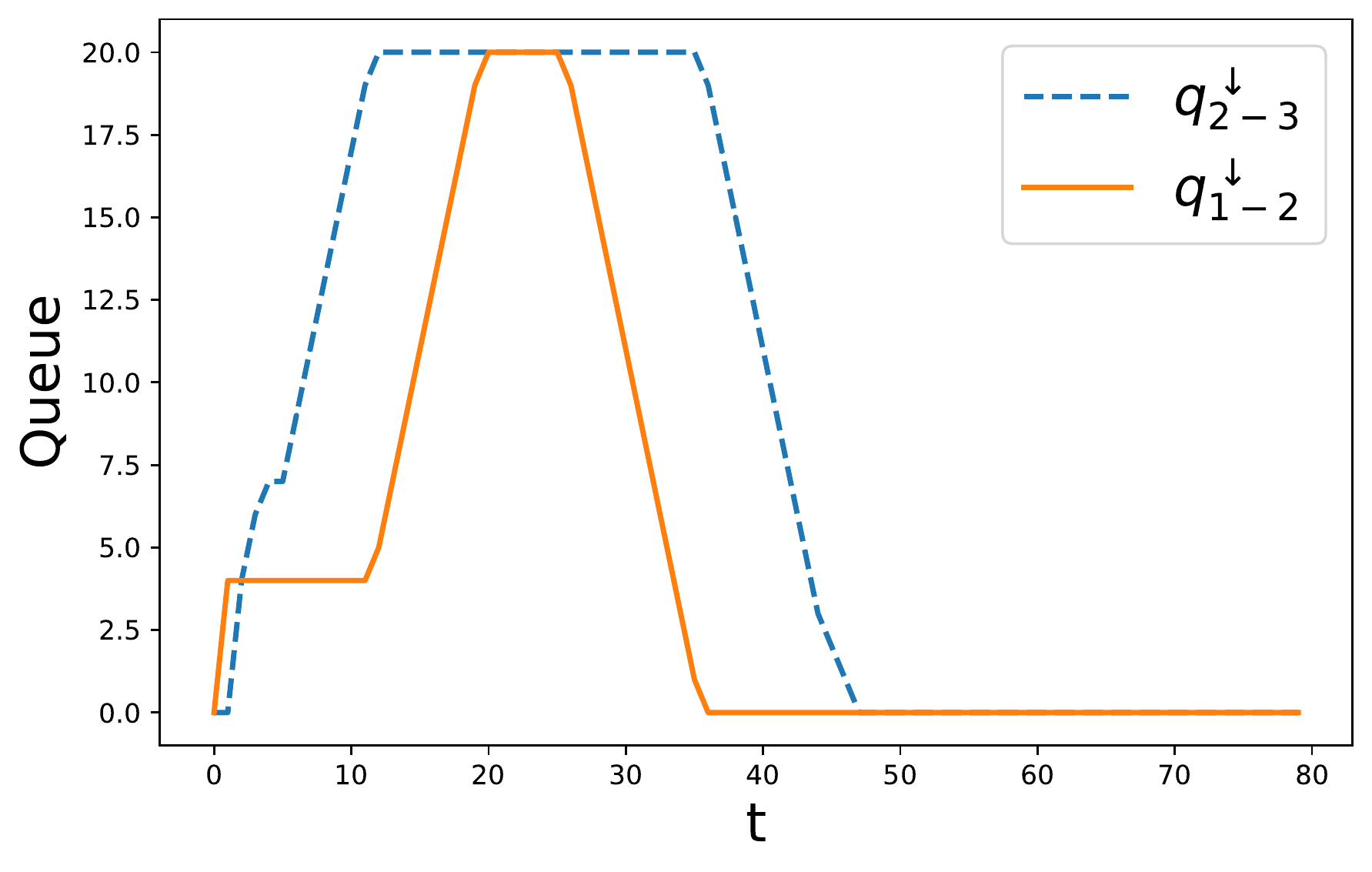}\label{subfig:queue_spill}~
	\caption{Queue spillback on the network for DUE}
	\label{fig:queue} 
\end{figure}

\end{exmp}

\begin{exmp}\label{due_ow}
In this example, we apply the developed algorithm to the OW network (13 nodes and 24 links) based on the LTM environment. The free flow speed is $v=0.2$ and the backward speed is $w=0.1$. Other parameters of the OW network is shown in Figure \ref{fig:ow_net}. The origin is 1 and the destination is 13. The travel demand is $d(0)=20$.

% \textcolor{blue}{where $q$ is flow rate, $\rho$ is link density, $\rho_{jam}$ is jam density and $u$ is speed. Note that the iterative method does not work well on this case, a nonlinear complementarity problem (NCP) can be formulated to solve DUE \textcolor{red}{(citation)}.}

Convergence plots of our algorithm and the traditional iterative method are presented in Figure~\ref{fig:ow_due}. In Figure \ref{subfig:ow_due_ite}, the average travel time converges to $5.5$ after 40 episodes. At the equilibrium, there are $9$, $1$, $4$ and $6$ vehicles choosing paths $1-2-8-9-13$, $1-7-8-9-13$, $1-7-8-12-13$ and $1-7-11-12-13$, respectively. For the MF-MA-DQL algorithm, after bouncing back and forth during the first $300$ episodes in Figure~(\ref{subfig:ow_due_ma}), average travel travel time gradually reaches its converged value around $5.5$, which is consistent with the iterative method.

\begin{figure}[H]
	\centering 
	\includegraphics[scale=.43]{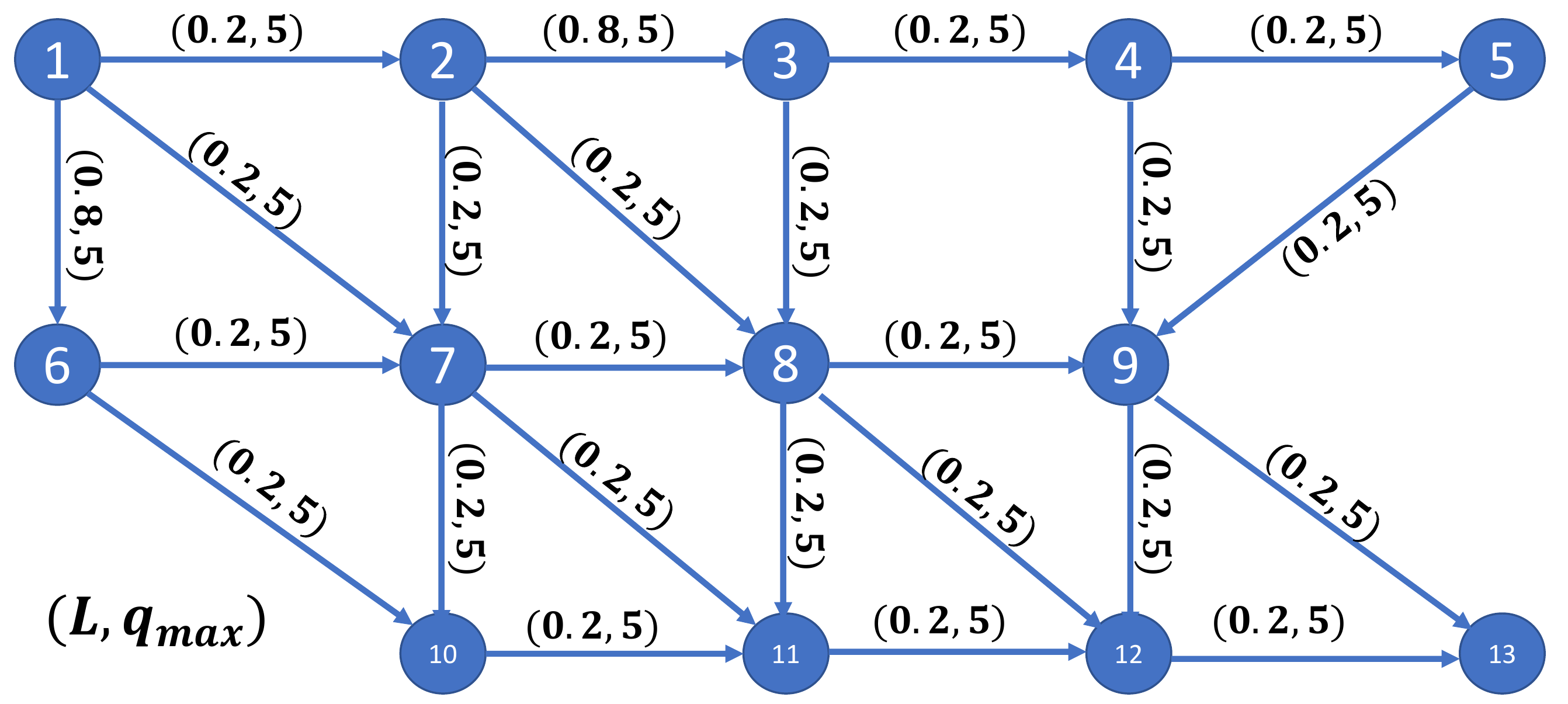}
	\caption{OW network}
	\label{fig:ow_net} 
\end{figure}

\begin{figure}[H]
	\centering 
	\subfloat[The iterative method]{\includegraphics[scale=.43]{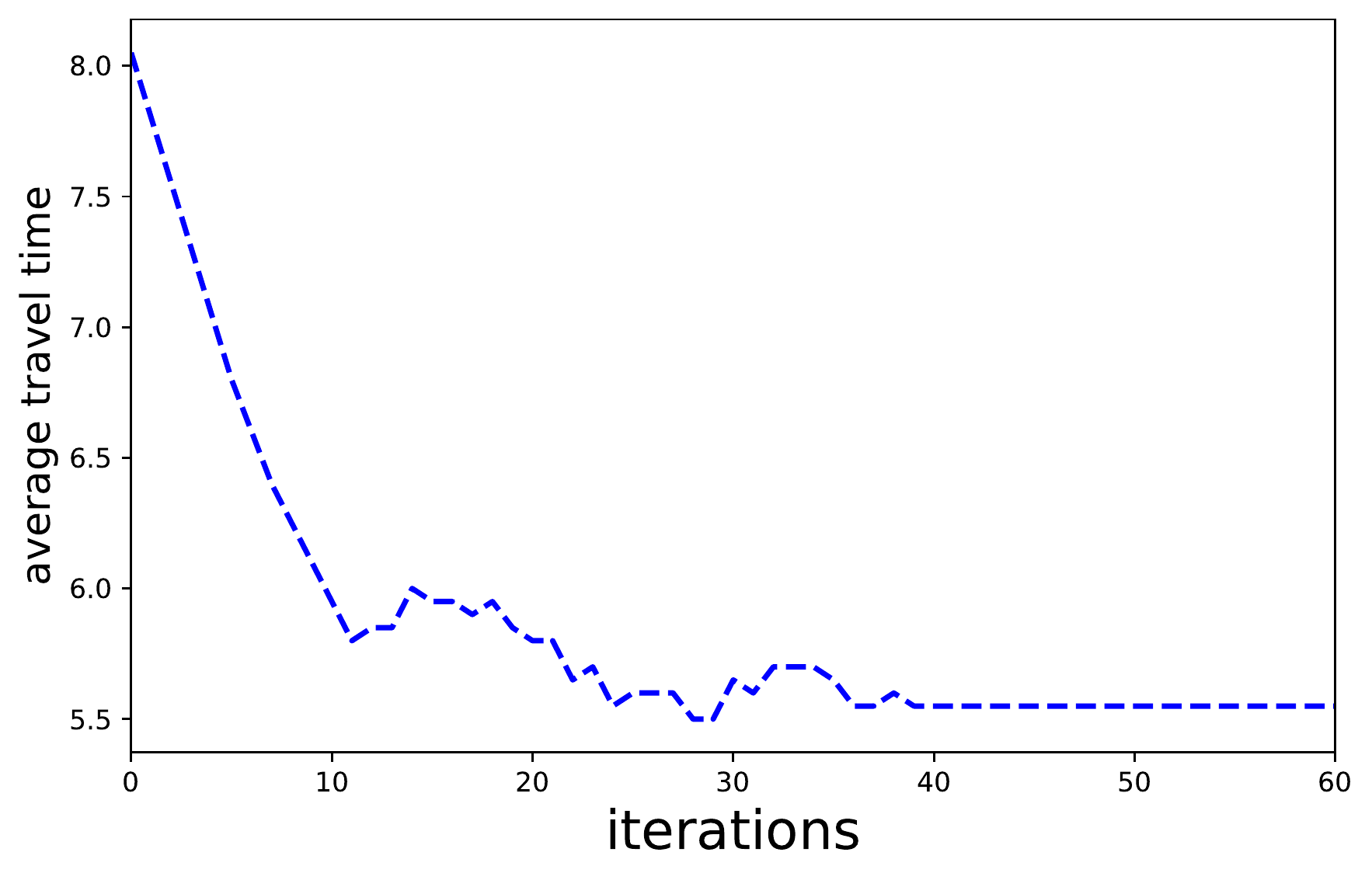}\label{subfig:ow_due_ite}}~~~
	\subfloat[MF-MA-DQL algorithm for DUE]{\includegraphics[scale=.43]{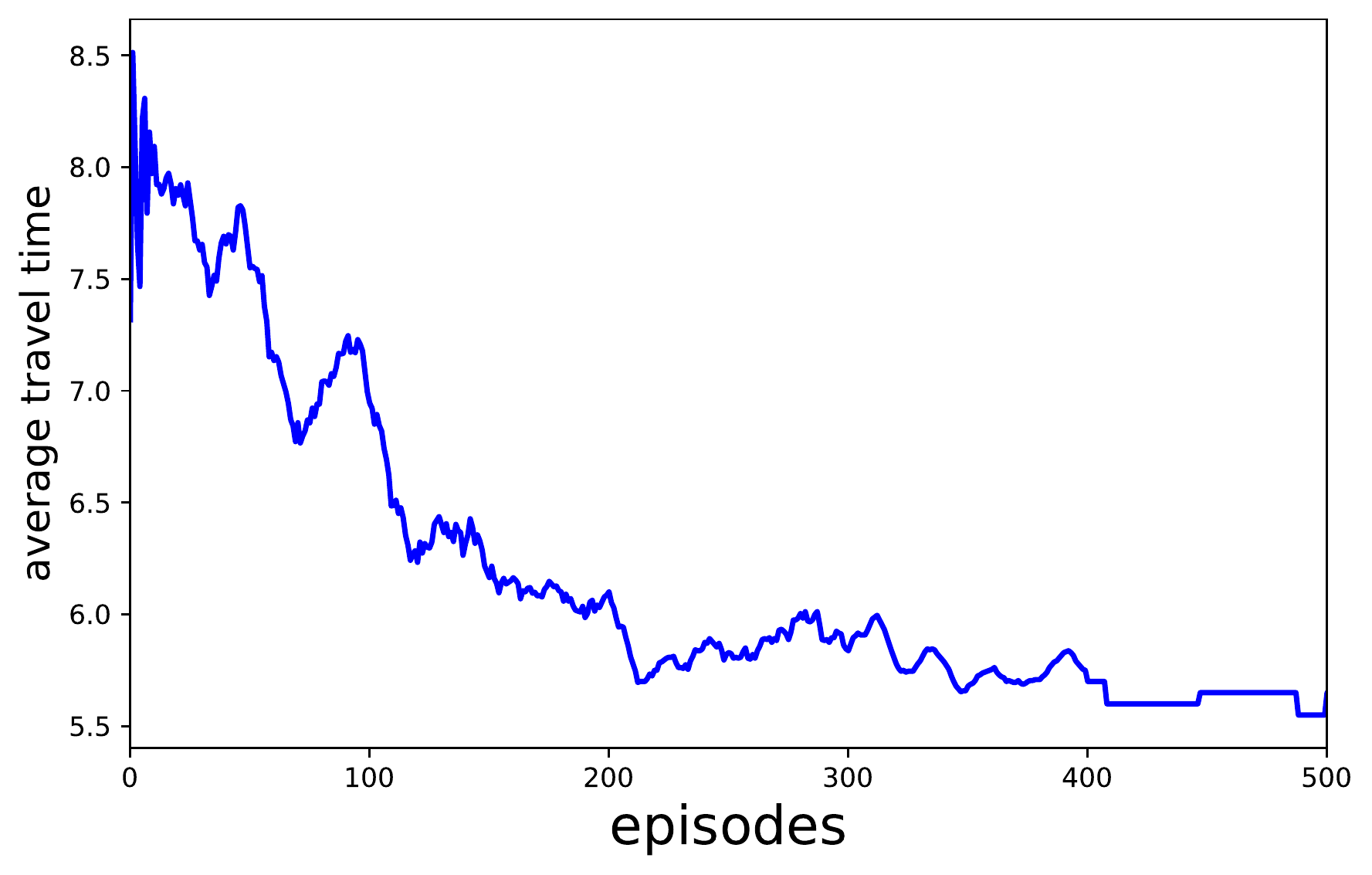}\label{subfig:ow_due_ma}}
	\caption{Convergence plots for DUE}
	\label{fig:ow_due} 
\end{figure}

\end{exmp}

\begin{exmp}\label{due_sumo} 
In this example, we apply the developed algorithm to a real-world road network with 69 nodes and 166 links, as presented in Figure~(\ref{fig:sumo_net}). This road network covers the area from 133th Street (south) to 146th Street (north) and from Riverside Drive (west) to Convent Avenue (east) in upper Manhattan. In addition to Riverside Drive and Convent Avenue, there are Broadway and Amsterdam Avenue in the north-south direction. The road network is imported into SUMO. 

\begin{figure}[H]
	\centering
	\includegraphics[width=0.99\linewidth,height=0.5\textheight,keepaspectratio]{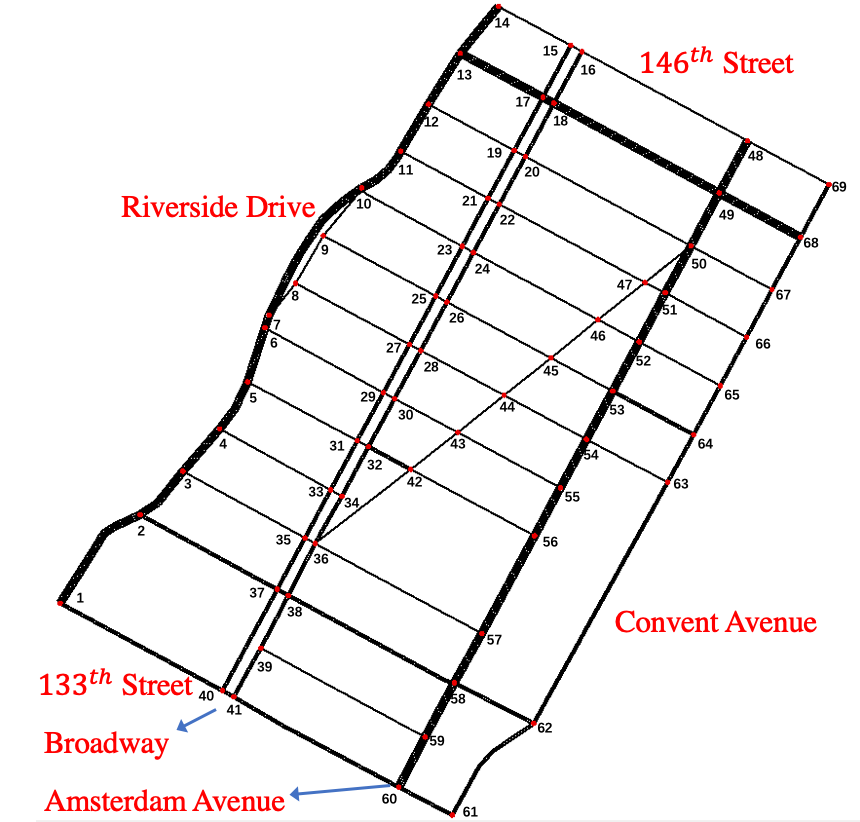}
	\centering 
	\caption{Road network in SUMO}
	\label{fig:sumo_net}
\end{figure}
%\tcr{I have updated this figure and uploaded, dont why it's still the old one. corrected it. please make sure the figure covers the area from 133th Street to 146th Street before submission. }

We first briefly introduce the environment. There are two types of vehicles in the road network, namely controllable agents and background traffic. The former actively adapts her route choices while the latter constantly follows some prescribed travel pattern. %In this case study, there are four groups of controllable agents: 1) three agents travel from node 14 to node 60, 2) three agents from node 15 to node 60, 3) three agents from node 48 to node 1, and 4) three agents from node 69 to node $1$. 
There are four groups of controllable agents in this case study, namely the first group travelling from node 14 to node 60, the second group from node 15 to node 60, third group from node 48 to node 1, and the fourth group from node 69 to node 1. In total there are 120 controllable agents who enter the road network according to their starting time. On average, there are 12 controllable agents entering the road network every 50 seconds. With respect to the background traffic, there are in total around 1,600 vehicles in the south-north direction and 500 vehicles in the east-west direction within the simulation time period (i.e., 1000 seconds). 

The goal of each controllable agent is to minimize her travel cost from origin to destination. Noticing that the first two groups of controllable agents share the same destination, i.e., node 60, these agents thus share the same $Q$ function. We denote this $Q$ function as $Q^{60}$, where the superscript $60$ indicates that this $Q$ function is used by agents whose destination is node $60$. Obviously, $Q$ value at destination node is $0$, i.e., $Q^{60}(o_i = (60,t),a_i,\bar{a}_i) = 0$, regardless of time $t$, action $a_i$, and mean action $\bar{a}_i$. Similarly, the remaining controllable agents whose destination is node $1$ share the same $Q$ function, which is denoted by $Q^1$. $Q^1(o_j=(1,t),a_j,\bar{a}_j) = 0$.

%With the aforementioned setup, we run the bilevel optimization model until convergence. 

% \tcb{We now apply the MF-MA-DQL algorithm to the aforementioned setup (i.e., 120 controllable agents and thousands of vehicles as background) in a default traffic scenario in SUMO (e.g., no traffic accidents). } \tcr{Is it appropriate to say default traffic scenario in Sumo? } \tcb{A convergence plot is presented in Figure~(\ref{fig:conv_BO}). The y-axis is the average travel time of all controllable agents and the x-axis is the index of episodes in training. Initially, agents spend a very long time (i.e., more than 950 seconds) on traveling to their destinations. Actually, some agents may not be able to reach their destinations and could record a total travel time of 1,000 seconds (i.e., the maximum allowed time in one episode). During the first 50 episodes, agents explore the road network and learn towards a better policy pretty fast. Therefore, their average travel time is substantially decreased from above 950 seconds to around 750 seconds. Despite some bouncing back and forth between 50 episodes and 150 episodes, the average travel time of all controllable agents stabilizes around 700 seconds after 150 episodes.}

% \begin{figure}[H]
% 	\centering 
%     \includegraphics[scale=.7]{./figures/sumo_120ag_300episods}\label{subfig:SUMO_BO_lower}~~~
% 	\caption{Convergence of MF-MA-DQL algorithm in the default SUMO setting}
% 	\label{fig:conv_BO} 
% \end{figure}	

\begin{figure}[H]
	\centering 
    \includegraphics[scale=.7]{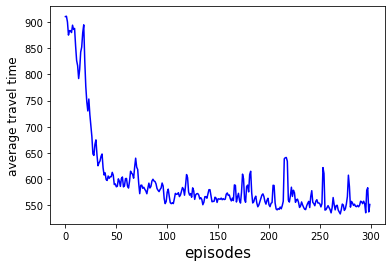}
	\caption{MF-MA-DQL algorithm (accident occurs)}
	\label{fig:block_converge} 
\end{figure}

To demonstrate the effectiveness of the proposed MF-MA-DQL algorithm, we now apply it to the aforementioned setup (i.e., 120 controllable agents and thousands of vehicles as background) with an accident blocking the link connecting node 26 and node 27, starting at $t=180 s$ and ending at $t = 490 s$. The convergence plot is presented in Figure~(\ref{fig:block_converge}). The y-axis is the average travel time of all controllable agents and the x-axis is the index of episodes in training. Initially, agents spend a very long time (i.e., more than 900 seconds) on traveling to their destinations. Actually, some agents may not be able to reach their destinations and could record a total travel time of 1,000 seconds (i.e., the maximum allowed time in one episode). During the first 50 episodes, agents explore the road network and learn towards a better policy pretty fast. Therefore, their average travel time is substantially decreased from above 900 seconds to around 600 seconds. Despite some bouncing back and forth between 50 episodes and 300 episodes, the average travel time of all controllable agents stabilizes around 570 seconds after 100 episodes.

Given that there is an accident happening from $t=180 s$ and lasting for more than 300 seconds, one natural question arises, i.e.,  whether agents' route choice behavior is changed by this accident. Therefore, we visualize agents' route choices before, during, and around the end of the accident in Figure~(\ref{fig:block_path}). At $t=80 s$, i.e., before the accident, controllable agents going to node 1 mainly choose Broadway instead of other north-south roads (i.e., Riverside drive, Amsterdam avenue, and Convent avenue) due to a comparatively light traffic condition on Broadway. At $t=190 s$, an accident occurs on the link connecting node 25 and node 27, and thus no vehicle could reach node 27 from node 25. In other words, no vehicle could go south by entirely staying on Broadway. As one may see in the middle plot of Figure~(\ref{fig:block_path}), controllable agents have learned the accident and are able to avoid being stuck in traffic by using the adjacent link as a slight detour. The route choice of agents at $t=440 s$ (i.e., around the end of the accident) show that some controllable agents are now willing to wait in traffic until the traffic accident is resolved, indicating that from the perspective of these agents, detouring may be more costly than waiting in traffic when they have learned that the accident is about to end.

%The adapting route choice behavior of controllable agents before, during, and around the end of the accident showcases that the proposed MF-MA-DQL mo 

\begin{figure}[H]
	\centering 
    \includegraphics[scale=.45]{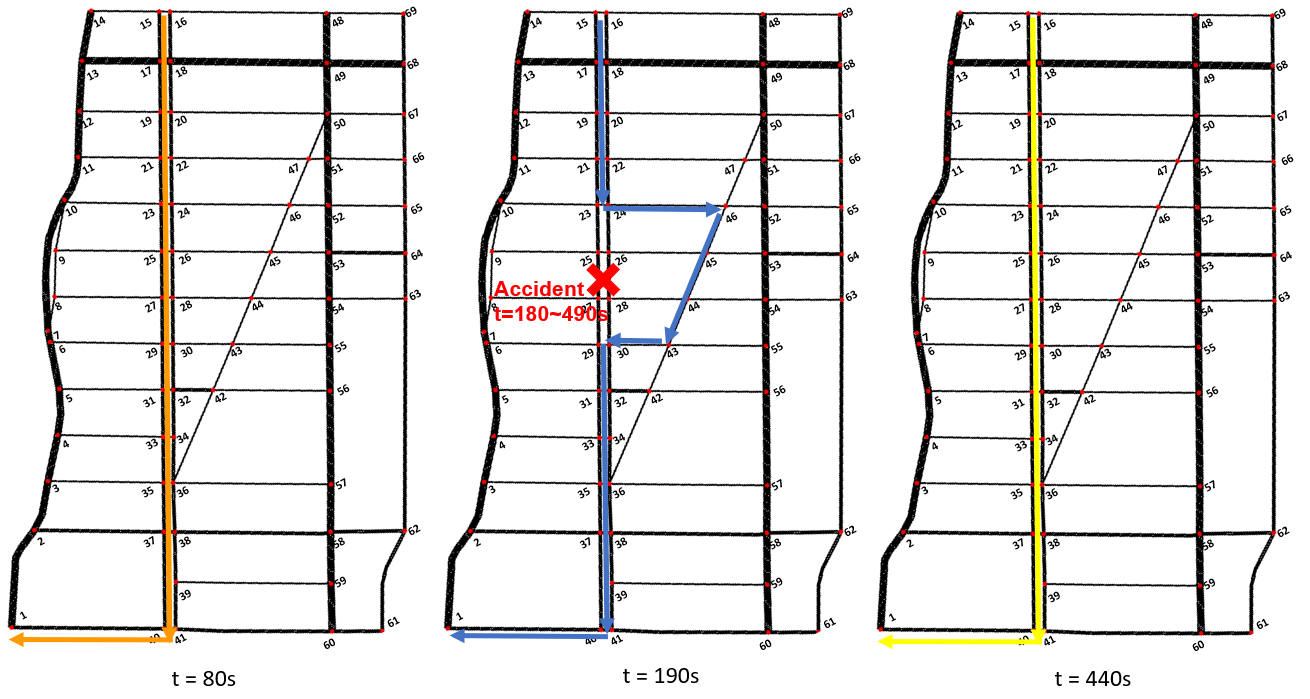}
	\caption{Route choice of controllable agents at different time steps}
	\label{fig:block_path} 
\end{figure}

%\textcolor{red}{We then consider another scenario based on our algorithm. So to verify the effectiveness of our algorithm, we built an example to simulate the occurrence of an accident. In this example, two vehicles will have a collision on Broadway, and the entire road from north to south will be blocked. The traffic accident handling time is set to be 300 seconds. And then traffic will resume on Broadway. All other settings are the same as the former one. We train our controllable agents to see their routing behavior in this dynamic environment.}

\end{exmp}

We list the computation time of all numerical examples in Table \ref{table:computation_time}. In Example 4.1-4.3, the computation time of the MARL approach is the running time of the training process until agents stabilize their behaviors. The computation time of the fixed point method is the running time of the iterative process until traffic flow reaches a stationary point. It is shown that the computation time of solving the Markov game is longer than that of the fixed point method. The explanation is in the Markov game, each individual agent needs to learn from the traffic environment and update their own policies. In Example 4.4, the computation time is the running time of the training process with traffic simulated in SUMO. Note that the computation time in this example is much longer than others. This is because the simulation is time-consuming. For each episode, the running time of the simulator is set as 20 min.  The total time of simulation is $300 \cdot 20 =10 h$. 

\begin{table}[H]
	\centering
	\begin{tabular}{c|c|c}
		\hline & MARL & Fixed point \\ \hline
		\makecell{4.1 Simple network \\ (50 agents, no spillback)} & 1 min 29s & 11s\\ 
		\hline \makecell{4.2 Simple network \\ (90 agents, spillback)} & 6 min 37s & 57s\\ \hline \makecell{4.3 OW network \\ (20 agents)} & 2 min 7s & 19s\\ \hline 
		\makecell{4.4 SUMO simulation \\ (120 agents, \\ 2100 background vehicles)} & \makecell{12h \\ (Simulation: 10h)} & ---\\ \hline
	\end{tabular}
	\caption{\small{Computation time}}
	\label{table:computation_time}
\end{table}

\section{Conclusion}\label{sec:conclu}

	 This paper develops a Markov game and multi-agent reinforcement learning that models intelligent agents' learning behavior and the system's equilibrating processes in a routing game among atomic selfish agents. 
	 A multi-agent mean field deep Q-learning algorithm is developed to tackle the route choice task of multiple self-interested agents. The proposed mean action, which is defined as the traffic flow on the chosen link, carries partial but not full information of nearby agents and is thus able to partially capture the competition among agents. In addition, the use of the mean action enables Q-value sharing and policy sharing, which is computationally efficient. The developed algorithm is demonstrated in four numerical examples, namely a simple network with various DNL models as its traffic propagation mechanism, the OW network with LTM, and a large-sized real-world road network with 69 nodes and 166 links implemented in SUMO. %The thorough testing unveils the versatility and effectiveness of the developed model-free MARL algorithm. 

	The linkage between the classic DUE paradigm and our proposed MARL paradigm is demonstrated schematically. 
	The difference is that, DTA models stipulate the system evolution dynamics and the travel cost of using each link or route, while the routing game does not contain such information and thus requires agents to learn and adapt their strategy according to historical episodes generated from interactions. 
	Specifically, the developed Markov routing game depicts DUE assuming perfect information and deterministic environments propagated by DNL models.
	On the simple network without and with spillback, we further show that the numerical solution from the developed MARL algorithm agrees well with the DUE from the iterative method. %To our knowledge, this paper is the first-of-its-kind to unify the mode-based (i.e., DUE) and data-driven (i.e., MARL) paradigms for dynamic routing games. 

    Nevertheless, there are several future directions we would like to explore. 
    % First, we assume that all travelers are perfectly rational in this study, meaning that travelers would always take the route with minimal expected travel time. However, 
    % the assumption of rationality has been challenged by various behavioral and economic theories, including, to name a few, bounded rationality \citep{di_boundedly_2013,di_boundedly_2016}, prospect theory \citep{gao2010adaptive,xu2011prospect,yang2014development,zhang2018cumulative}, and regret theory \citep{de2011expected}. 
    % %In particular, bounded rationality \citep{jayakrishnan_evaluation_1994, di_boundedly_2013} suggests travelers may not switch from their current route to an alternative route with less travel cost if the difference is not large enough. 
    First, %in addition to route choice, 
    departure time choice and velocity control \citep{huang2021driving} are two travel choices of intelligent agents. 
    %We leave bounded rationality and the simultaneous route and departure time choice in future research. 
    %We will also model how agents optimally select both velocity control and route choice, which requires agents to learn an equilibrium that involves both continuous and discrete controls \citep{huang2021driving}. 
    The proposed model-free MARL paradigm will be extended to accommodate departure time choice and velocity control for discrete travelers. Second, in this paper, we assume drivers make decisions when they reach a node. In reality, drivers may make routing choices before they arrive at a node. We would like to incorporate such behavior in future work. 
    Last but not the least, we will develop a bi-level network design problem that optimizes planning or policy countermeasures in the upper level and the adaptive routing behavior of discrete travelers in the lower level.

\section*{Acknowledgements}
This work is partially sponsored by the Region 2 University Transportation Research Center (UTRC) (subcontract RUTGER PO 966112/PID\#824227) and the National Science Foundation under CAREER award number CMMI-1943998.

%\section*{Reference}
\bibliographystyle{elsarticle-harv}
\bibliography{routeChoice}
%\end{multicols}
\appendix
\section{Appendices}

\subsection{}
\label{append:DNL_environment}

\textbf{PQ Environment:} The sending flow in the PQ environment is calculated in the same way as LTM. Different from the LTM environment, PQ does not capture the link capacity. Therefore, the receiving flow is only determined by the flow rate capacity:
\begin{align}
    R(t)= q_{max} \Delta t. \nonumber 
\end{align}

\textbf{SQ Environment:} The SQ model takes the link capacity into consideration. Different from the LTM environment, it does not capture the backward speed. The receiving flow is calculated as:
\begin{align}
    R(t)=min\{N^{\downarrow}(t)+k_j L-N^{\uparrow}(t),\ q_{max} \Delta t \}, \nonumber 
\end{align}

\textbf{CTM Environment:} In the CTM environment, each link is divided into cells. The length of each cell is supposed to be the distance traveled by a vehicle at free flow speed $v$. The sending flow $S(t)$ is the number of vehicles which can leave a cell at each time step. It is calculated as:
\begin{align}
    S(t)=min \{n(t), q_{max} \Delta t \}, \nonumber
\end{align}
where $n(t)$ is the number of vehicles on the cell at time $t$. It means the number of vehicles which can leave may not exceed the number of vehicles on the cell. The receiving flow $R(t)$ is the number of vehicles which can be accommodated on the cell at time $t$. It is calculated as:
\begin{align}
    R(t)=min \{\frac{w}{v}(N-n(t)), q_{max} \Delta t \}, \nonumber
\end{align}
where $N$ is the cell capacity, representing the maximum number of vehicles on a cell.

\subsection{}
\label{append:iteration_method}
\textbf{Iterative Method:} We employ the iterative method proposed in \cite{gawron_iterative_1998} to solve DUE. The basic idea of the iterative method is to let vehicles choose routes according to some policy and then calculate the network loading and travel cost by simulation so that vehicles gradually learn towards better route choices. By iterating this process, a stationary fixed point solution is supposed to be derived so that no vehicle could get better off by switching her route choices. 
To make this paper self-explanatory, we now briefly introduce the iterative method adapted to the two-node two-link network. Interested readers could refer to \cite{gawron_iterative_1998} for more details. 

\begin{algorithm}[H]
	\caption{An iterative method to solve DUE}
	\label{alg:iterative}
	\begin{algorithmic}[1]
		\State Initialize a proportion $p_t(0)$ indicating the proportion of the demand at $t$ choosing link $0$. Consequently, $p_t(1) = 1 - p_t(0)$
		\State Initialize a learning step size $\eta \in (0,1]$ 
		\State Initialize function $g(x) = exp(\dfrac{ax}{1-x^2})$, where $a$ is a parameter, and function $f(x) = \dfrac{p_t(0)g(x)}{p_t(0)g(x) + p_t(1)}$
		\Repeat
		\State Calculate the time-dependent travel cost of both links, denoted by $\tau_t(0)$ and $\tau_t(1)$, from simulation
		\State Calculate the relative cost difference $\delta_{01} = \dfrac{\tau_t(1) - \tau_t(0)}{\tau_t(1) + \tau_t(0)}$
		\State Update the proportion by $p_t(0) = (1-\eta) p_t(0) + \eta f(\delta_{01})$
		\State Calculate $p_t(1) = 1 - p_t(0)$
		\State Update function $f(x) = \dfrac{p_t(0)g(x)}{p_t(0)g(x) + p_t(1)}$
		\Until{the iterative method reaches a stationary fixed point}
		\State Return $p_t(0)$ and $p_t(1)$
	\end{algorithmic}
\end{algorithm} 

The iterative method is listed in Algorithm~(\ref{alg:iterative}). With demand $d(t)$ emerging from node $O$ at time $t$, we randomly initialize $p_t(0)$ and $p_t(1)$ to denote the proportion of the demand choosing link $0$ and link $1$, respectively. We also initialize functions $g(x)$ and $f(x)$ which will be used to update $p_t(0)$ and $p_t(1)$. Note that $f:[-1,1]\rightarrow [0,1]$ is a monotonic increasing function. With $p_t(0)$ and $p_t(1)$, we run simulation to calculate the average travel cost on both links for the demand starting at time $t$ and denote these time-dependent travel cost by $\tau_t(0)$ and $\tau_t(1)$. We then calculate the relative cost difference $\delta_{01} = \dfrac{\tau_t(1) - \tau_t(0)}{\tau_t(1) + \tau_t(0)}$ and use this cost difference to update the proportion $p_t(0)$ towards $f(\delta_{01})$ by step size $\eta$. The rationale of this updating rule is as follows. When $\delta_{01}$ is large (e.g., close to $1$), meaning that the cost on link $0$ is much less than that on link $1$, we have $f(\delta_{01}) \approx 1$, indicating that $p_t(0)$ is updated towards a larger value; when $\delta_{01}$ is small (e.g., close to $-1$), meaning that the cost on link $0$ is much larger than that on link $1$, we have $f(\delta_{01}) \approx 0$, indicating that $p_t(0)$ is updated towards a smaller value. After updating $p_t(0)$ and $p_t(1)$, we repeat the iterative process until reaching a stationary fixed point. 

%	\paragraph{Comparison between the solution of the iterative method and MF-MA-DQL solution}
%	\begin{figure}[H]
%		\centering
%		\includegraphics[width=0.99\linewidth,height=0.25\textheight,keepaspectratio]{./figures/due}
%		\centering 
%		\caption{Convergence plot of the MF-MA-DQL algorithm for DUE}
%		\label{fig:due}
%	\end{figure}
\end{document}